\documentclass{article}

\usepackage{arxiv}

\usepackage[utf8]{inputenc} 
\usepackage[T1]{fontenc}    
\usepackage[table,xcdraw,svgnames,usenames,dvipsnames]{xcolor}

\usepackage[colorlinks]{hyperref}
\AtBeginDocument{%
  \hypersetup{
    citecolor=RawSienna,
    linkcolor=Red,   
    urlcolor=Black}}
\usepackage[numbers]{natbib}
\usepackage{enumitem,kantlipsum}
\usepackage{multicol}
\usepackage{url}            
\usepackage{booktabs}       
\usepackage{amsfonts}       
\usepackage{nicefrac}       
\usepackage{microtype}      
\usepackage{lipsum}		
\usepackage{graphicx}
\usepackage[english]{babel}
\usepackage{doi}
\usepackage{multirow}
\usepackage{amsmath}
\usepackage{svg}
\usepackage[normalem]{ulem}
\usepackage{natbib}
\setcitestyle{square, comma, numbers,sort&compress}

\def\code#1{\texttt{#1}}
\usepackage{ffcode}
\usepackage{tcolorbox}
\usepackage{titlesec}

\usepackage{cleveref}

\usepackage[toc,page,header]{appendix}
\usepackage{minitoc}

\renewcommand \partname{}

\definecolor{myblue}{RGB}{0,163,243}
\tcbuselibrary{skins,xparse,breakable}
\tcbset{mystyle/.style={
  breakable,
  enhanced,
  outer arc=0pt,
  arc=0pt,
  colframe=myblue,
  colback=myblue!20,
  attach boxed title to top left,
  boxed title style={
    colback=myblue,
    outer arc=0pt,
    arc=0pt,
    top=3pt,
    bottom=3pt,
    },
  fonttitle=\sffamily
  }
}

\tcbset{mystyle0/.style={
  breakable,
  fonttitle=\sffamily
  }
}

\newtcolorbox[auto counter,number within=section]{prompt0}[2][]{
  mystyle0,label=#2,
  title=Prompt~\thetcbcounter
}

\newtcolorbox[auto counter,number within=section]{promptOLD}[2][]{
  label=#1,
  title={Prompt \thetcbcounter \quad  #2}
}

\newtcolorbox[auto counter,number freestyle={\noexpand Prompt \noexpand\arabic{\tcbcounter}}]{prompt}[2][]{
  label = #1,
  title={\thetcbcounter \quad  #2}
}

\newtcolorbox[auto counter,number freestyle={\noexpand Completion \noexpand\arabic{\tcbcounter}}]{completion}[2][]{
  label = #1,
  title={\thetcbcounter \quad  #2}
}

\newtcolorbox[auto counter,number freestyle={\noexpand Argument \noexpand\arabic{\tcbcounter}}]{argument}[2][]{
  label = #1,
  title={\thetcbcounter \quad   #2}
}

\newtcolorbox[auto counter,number freestyle={\noexpand Story \noexpand\arabic{\tcbcounter}}]{story_example}[2][]{
  label=#1, 
  title={Example \thetcbcounter \quad   #2},breakable, 
  coltitle = black,
  colbacktitle=gray!5!white,
  colback=gray!1!white
}

\newtcolorbox[auto counter,number freestyle={\noexpand Example \noexpand\arabic{\tcbcounter}}]{example}[2][]{
  label=#1,
  title={\thetcbcounter \quad   #2},breakable, 
  coltitle = black,
  colbacktitle=gray!5!white,
  colback=gray!1!white
}

\newcommand{\gray}[1]{ {\color{gray} #1} }

\title{Large-scale study of human memory for meaningful narratives}

\author{ \href{https://orcid.org/0000-0003-3962-5836}{\includegraphics[scale=0.06]{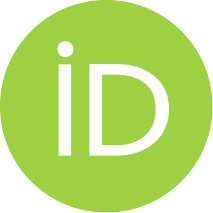}\hspace{1mm}Antonios~Georgiou}
\thanks{ Equal contribution } \\
School of Natural Sciences\\
	Institute for Advanced Study\\
	Princeton, NJ \\
	\texttt{antoine.georgiou@gmail.com } \\
	\And
	\href{https://orcid.org/0000-0002-0999-2355}{\includegraphics[scale=0.06]{orcid.pdf}\hspace{1mm}Tankut~Can} 
 \footnotemark[1] \\
	Department of Physics\\
	Emory University\\
	Atlanta, GA \\
	\texttt{tankut.can@gmail.com } \\
 	\And
	\href{https://orcid.org/0000-0002-9532-8484}{\includegraphics[scale=0.06]{orcid.pdf}\hspace{1mm}Mikhail Katkov} \\
	Department of Brain Sciences\\
	Weizmann Institute of Science\\
	Rehovot, 76100 Israel \\
        School of Natural Sciences\\
	Institute for Advanced Study\\
	Princeton, NJ \\
	\texttt{mikhail.katkov@gmail.com} \\
  	\And
	\href{https://orcid.org/0000-0002-5661-4349}{\includegraphics[scale=0.06]{orcid.pdf}\hspace{1mm}Misha Tsodyks}\thanks{Corresponding author} \\
	Department of Brain Sciences\\
	Weizmann Institute of Science\\
	Rehovot, 76100 Israel \\
        School of Natural Sciences\\
	Institute for Advanced Study\\
	Princeton, NJ \\        
	\texttt{mtsodyks@gmail.com} \\
}

\renewcommand{\shorttitle}{Narrative Memory}

\hypersetup{
pdftitle={using large language models to study human memory for meaningful narratives},
pdfsubject={q-bio.NC, q-bio.QM,cs.AI,cs.LG},
pdfauthor={Antonios~Georgiou, Tankut~Can, Mikhail~Katkov, Misha~Tsodyks},
pdfkeywords={large-scale study, text segmentation, free recall, recognition memory, clauses, prompt engineering}
}

\begin{document}

\doparttoc 
\faketableofcontents 


\maketitle

\begin{abstract}
The statistical study of human memory requires large-scale experiments, involving many stimuli conditions and test subjects. While this approach has proven to be quite fruitful for meaningless material such as random lists of words, naturalistic stimuli, like narratives, have until now resisted such a large-scale study, due to the quantity of manual labor required to design and analyze such experiments. In this work, we develop a pipeline that uses large language models (LLMs) both to design naturalistic narrative stimuli for large-scale recall and recognition memory experiments, as well as to analyze the results. We performed online memory experiments with a large number of participants and collected recognition and recall data for narratives of different sizes. We found that both recall and recognition performance scale linearly with narrative length; however, for longer narratives people tend to summarize the content rather than recalling precise details. To investigate the role of narrative comprehension in memory, we repeated these experiments using scrambled versions of the narratives. Although recall performance declined significantly, recognition remained largely unaffected. Recalls in this condition seem to follow the original narrative order rather than the actual scrambled presentation, pointing to a contextual reconstruction of the story in memory. Finally, using LLM text embeddings, we construct a simple measure for each clause based on semantic similarity to the whole narrative, that shows a strong correlation with recall probability. Overall, our work demonstrates the power of LLMs in accessing new regimes in the study of human memory, as well as suggesting novel psychologically informed benchmarks for LLM performance. 

\end{abstract}


\section{Introduction}

In the classical paradigm for studying human memory, participants are presented with randomly assembled lists of words and then perform memory tasks such as recognition and recall (see review in \citep{kahana20review}). A wealth of results has been obtained in these studies. For instance, it has been found that words at the end and the beginning of the list have a higher chance of being recalled (recency and primacy effects, respectively), and there is a tendency to recall words close to each other in the list (contiguity, \citep{kahana1996associative}). Moreover, it was found that when the presented lists grow in length, even though the average number of recalled words ($R$) is increasing, a progressively smaller fraction of the words is recalled \citep{murdock1962serial}. Several authors have addressed the issue of the mathematical form of the dependence of $R$ on list length and found that the best description for this dependence is provided by power-law relations, $R \sim L^\alpha$, with exponents $\alpha$ generally below one \citep{murray1976standing}. It is well known that recall also depends on multiple experimental factors such as e.g. the presentation rate of words, the age of the participants, etc. However, in recent work, some of the authors discovered that if recall performance is analyzed as a function of a number of \textit{retained} ($M$), rather than presented words, the relation becomes universal and is described by the analytical form: $R = \sqrt{\frac{3 \pi}{2}M}$ \citep{naim2020fundamental}. Moreover, this relation follows from a simple deterministic model where words are retrieved one by one according to a random symmetric matrix of `similarities' reflecting their long-term encoding in memory, until the process enters a cycle and no more words can be recalled. The number of words retained in memory  $M$, itself can be predicted by the \textit{retrograde interference} model that assumes that each new word erases some of the previously presented words according to the `valence' or `importance' of each word \citep{georgiou2021retroactive,georgiou2023forgetting}. 

While it is remarkable that human memory for random material can be described with universal mathematical relations, it is of course much more important and exciting to try to understand how people remember more natural, meaningful information. After the pioneering work of \citet{bartlett}, many studies considered recalls of narratives. As opposed to random lists, narratives convey meaning, and hence have structure on multiple levels which influences recall, as was confirmed in many previous publications (see \Cref{sec:prev-work}). The first challenge in understanding narrative recall is the fact that people tend not to recall the narrative verbatim. Rather, they remember what the narrative is about and retell it in their own words \citep{Gomulicki1956,Fillenbaum1966,Sachs1967}. Counting correctly recalled words is therefore not a good score of recall, and the better score used in many studies, that we also adopt in our work, is a count of recalled `ideas', or `clauses' (see e.g. \citep{bransford1972contextual}). Using this method however requires a human-level of understanding of narratives and recalls, making collecting large amounts of data difficult and extremely time-consuming to analyze. In our study, we develop a way to overcome this and other difficulties by using large language models (LLMs) to  assist in the analysis and design of experiments, as described later. In particular, we use LLMs to generate new narratives of a particular type and length, and to score human recalls obtained in multiple experiments performed over the internet. In addition to recall, we also performed \textit{recognition} experiments (where people are requested to indicate whether a specific clause was in the presented story or not) in order to estimate how many clauses people remember after reading the narrative. To this end, we use LLMs to generate plausible lures, i.e. novel clauses that could have potentially appeared in the narrative. 

Inspired by our previous results, we wanted to understand how recognition and recall performance scale up with narrative length as it increases and what the relation between them is. To this end, we performed a large number of experiments over the internet using the Prolific platform (\url{www.prolific.com}). We also compared the recall and recognition performance of original narratives with their scrambled versions in order to elucidate the effects of comprehension on different aspects of memory. Since there are different types of narratives that could potentially be more or less difficult for people to remember and recall, we decided to focus on one particular type of narrative first studied in the famous paper by \citet{Labov1966} that established the field of narratology, namely the oral retelling of personal experience, told by real people, and the variants of those generated by LLMs (see later). While being collected in a research setting, these spoken recollections of dramatic personal episodes are close to the natural way people share their experiences in real life and therefore are of special interest for studying human memory.  


\section{Results}

\subsection{LLM-assisted recall and recognition experiments.}

For the purpose of this study, we have chosen several narratives of different lengths from \citet{labov2013language} and \citet{Labov1966}. As part of the analysis in these publications, narratives were segmented into an ordered set of \textit{clauses}, which are ``the smallest unit of linguistic expression which defines the functions of narrative'' \citep{Labov1966}. In other words, they are the smallest meaningful pieces which still serve some function in communicating a narrative. 
Since these are spontaneous narratives spoken in local dialect, they are characterized by a number of features which are awkward to transcribe (pauses, repetition, gestures) as well as non-standard (and sometimes outdated) English vernacular. These factors complicate comprehension when participants have to read narratives on the computer screen. We therefore instructed LLMs to generate new narratives modeled on the original ones, i.e. exhibiting a similar type of a event sequence and the overall length in terms of the total number of clauses. In particular, the LLM-generated narratives inherited the segmentation from the original story, i.e. the number of clauses was the same and the information contained in the corresponding clauses had a similar role in their respective narrative (see Methods \Cref{methods} for details of narrative generation, and some examples in Appendix \ref{sec:narr_gen}). Eight narratives were selected for subsequent memory experiments, ranging from 18 to 130 clauses in length. We presented each narrative to a large number of participants ($\sim200$) who then performed either recall or recognition tasks. In the subsequent analysis, we treated clauses as the basic units that together communicate the meaningful information contained in the narrative. In particular, we quantified each individual recall by identifying which of the clauses in the narrative were recalled, determining this by whether information contained in this clause is present in the recall. We simplify the analysis by considering each clause as being either recalled or not. This scoring of recalls is traditionally performed by human evaluators and is very time-consuming. We therefore \textit{prompted} an LLM to define which of the clauses of the original narrative were recalled and in which order. Here we utilized the remarkable ability of modern LLMs to respond to instructions, provided as {\it prompts} written in standard English (as opposed to a programming language), to perform novel tasks without any additional training (known as zero-shot prompting or in-context learning\citep{brown2020language}, see Appendix \ref{app:prompts_and+completions} for more details).

\begin{figure}[ht]
	\centering
 \includegraphics[width=\textwidth]{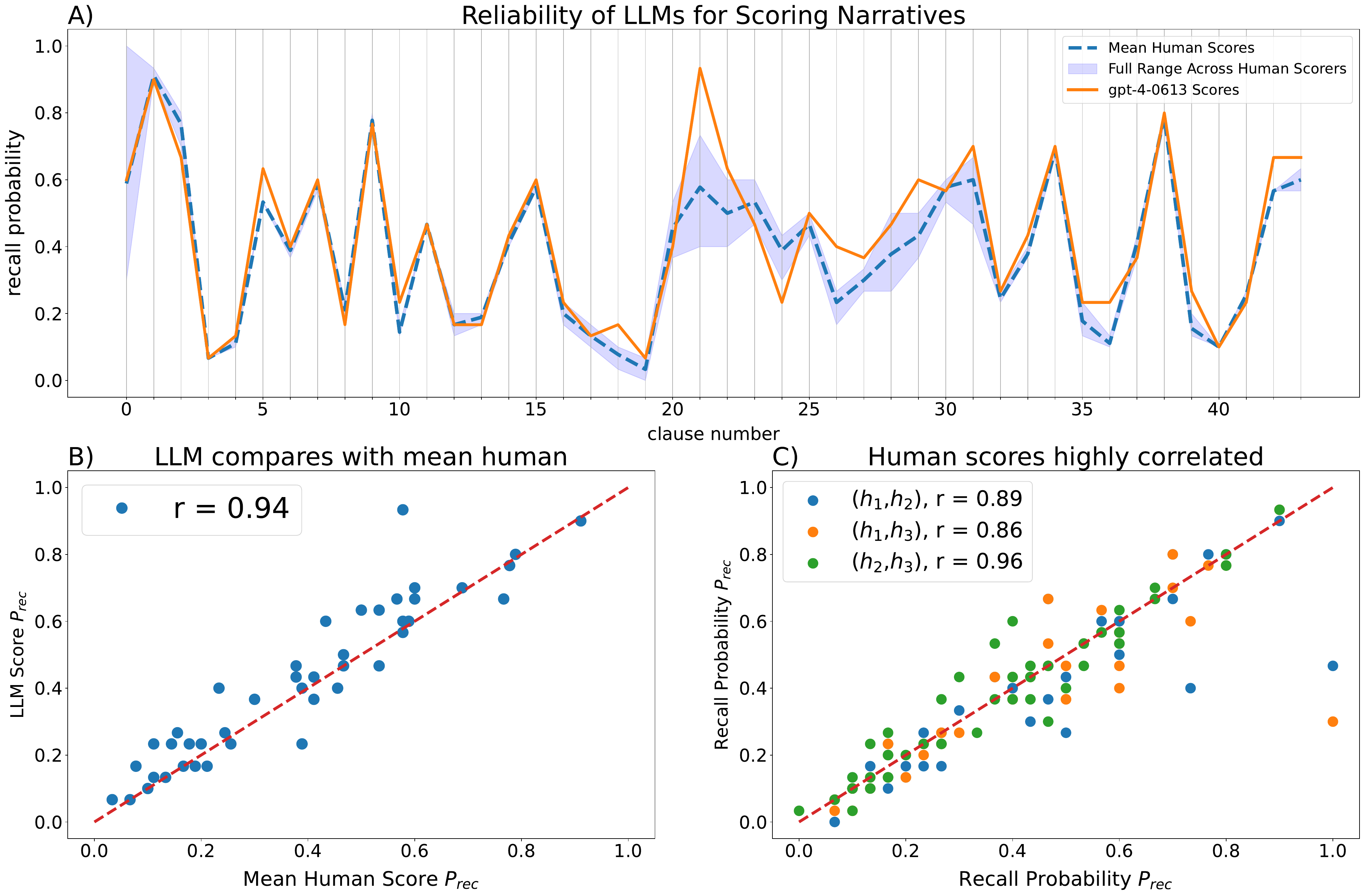}
	\caption{ {\it Reliability of LLM scoring of recalls}. 3 authors and GPT-4 (OpenAI API model \ff{gpt-4-0613}) performed scoring of 30 recalls by answering the question of whether the information present in each particular clause is present in each individual recall. (A): comparison between recall probabilities $P_{rec}$ for each clause, as calculated from GPT-4 scores (orange) and average human scores (dashed blue). The full range of human scores is given by the shaded blue region. (B): A strong correlation between human and GPT-4 scores across clauses, with a correlation coefficient (r-value) of 0.94. (C): Correlations between individual human scores, shows overall a strong agreement between human scorers.}
	\label{fig:reliability}
\end{figure}

To test the ability of the LLM to adequately score human recall (with appropriate prompting as described in the Methods \Cref{methods} and \Cref{app:prompts_and+completions}), we performed an additional set of recall experiments with a specially LLM-generated narrative and compared the LLM-performed recall scoring to the one conducted manually by the authors (see Methods \Cref{methods} for details). To this end, we calculated the fraction of participants who recalled each particular clause, i.e. the clause's recall probability ($P_{rec})$, as judged by the LLM and by the authors. For nearly all of our analysis, the LLM we used was OpenAI's GPT-4 (see \Cref{sec:different_LLM} for comparison between different LLMs). As shown in \Cref{fig:reliability}, GPT-4 scoring of recalls results in recall probabilities close to ones obtained by human evaluations for a great majority of the clauses. Moreover, variability of scoring is comparable between GPT-4 and human evaluators (compare \Cref{fig:reliability}B and C ). Interestingly, the LLM has a greater correlation with the mean human scoring  (r = 0.94) than with any individual scoring (r = 0.92, 0.90, 0.90) (see \Cref{table:scoring} in \Cref{sec:LLM_reliability}; c.f. \citet{Michelmann2023}).

We also performed recognition experiments in order to estimate the average number of clauses that participants remember after presentation. As we explain below, this analysis requires a large number of plausible \textit{lures}, i.e. novel clauses that could have possibly been in the narrative and hence cannot be easily distinguished from the true clauses using context and style. Generating these lures is highly nontrivial as it requires an understanding of the narrative. This makes manually generating lures very challenging and time-consuming, which is why we utilized LLMs for this purpose (see \Cref{sec:lures} for prompts and example output). Using the LLM, for each story we obtained the same number of lures as true clauses (denoted $L$ for a given story), and sampled 10 clauses from this entire pool of $2L$ clauses uniformly and randomly for the testing phase. Participants would then see one clause at a time, and were asked whether the clause was in the presented narrative or not. We checked that presenting several clauses for recognition does not result in systematic drift in performance, i.e. no output interference was detected in our experiments (\citet{criss2011output}; see Appendix \ref{app:output_interference}) .  We then estimated the number of clauses retained in memory after presentation of the narrative ($M$) from the fraction of `hits', i.e. correct recognitions of the true clauses ($P_{h}$) and the fraction of `false alarms', i.e. reporting lures as true clauses ($P_{f}$). We assume that if the participant remembers a given clause, they always recognize it as being part of the narrative; otherwise they still give a positive answer with some guessing probability $P_g$. The guessing probability is the probability that they respond ``yes" to a clause they did not retain, conditioned on having just read the narrative. We approximate this by the false alarm probability $P_{f}$, which is the probability that a participant responds ``yes" to a clause that {\it was not in} the story, also conditioned on having just read the story. For this approximation to make sense, the lures must appear as if they could have been in the story, thus making the performance on a lure appear as close as possible to performance on a true clause that was just not encoded.  Therefore, using $P_{g} \approx P_{f}$, the total probability of a correct recognition is given by

\begin{align} 
  P_h = \frac{M}{L} + \left(1-\frac{M}{L} \right) P_g \approx \frac{M}{L} + \left(1-\frac{M}{L} \right) P_f,
\end{align}
from which we obtain

\begin{equation}
    \frac{M}{L} = \frac{P_h - P_f}{1-P_f} .
    \label{recog}
\end{equation}
This equation emphasizes the importance of using lures in recognition experiments, since without them we have no way of estimating the ``guessing correction" to the hit probability.




\paragraph{Definition of Mathematical Variables }

Here for convenience, we provide a summary of the main mathematical variables used in our analysis:

\begin{itemize}
    \item $R$ - total number of clauses from a narrative recalled by an individual, averaged over all subjects.
     \item $L$ - total number of clauses in a narrative.
    \item $P_{rec}(c)$ - recall probability of clause $c$, determined by the number of subjects who recalled clause $c$, divided by the total number of subjects in the experiment. Often the argument is not explicitly written.
    \item $P_{h}$ - hit probability, computed from yes/no recognition experiments as the ratio of the total number of hits $H$ (over all subjects) divided by $H$ plus the total number of misses $M$, i.e. $P_{h} = H/(M + H)$.
    \item $P_{f}$ - false alarm probability, computed from yes/no recognition experiments as the ratio of the total number of false alarms $FA$ (i.e. false positive) divided by $FA$ plus the total number of correct rejections $CR$ (true negative), i.e. $P_{f} = FA/(FA + CR)$. 
    \item $M$ - estimate of the total number of clauses retained in memory, averaged over all subjects. This is estimated from $P_{h}$ and $P_{f}$ using \Cref{recog}.
\end{itemize}

\subsection{Scaling of Recall and Memory}

\begin{figure}[t]
 \includegraphics[width = \textwidth]{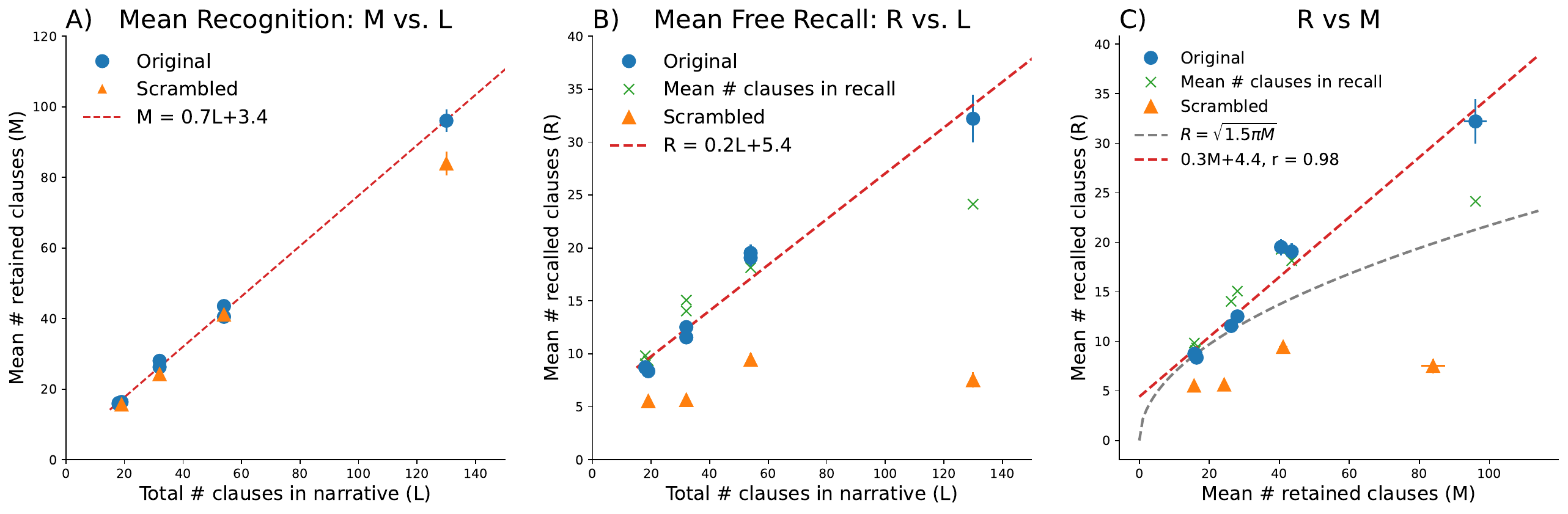}
	\caption{ {\it Human performance in recall and recognition experiments for narratives of different length}. (A): Estimated number of retained clauses (M) is plotted as a function of the number of clauses in the narrative (L) measured in recognition experiment. Surprisingly M has similar values in intact and scrambled narrative. (B): Average number of recalled clauses (R) for narratives of different length. In contrast to the M, R drops substantially for scrambled narratives. Also plotted are the average number of clauses used in the recall (green cross), which dips substantially below $R$ for longer narratives, indicating the tendency of subjects to summarize.  (C): Average number of recalled clauses vs. number of retained clauses from the same story. As expected from panels a) and b) the number of retrieved clauses in scrambled narrative is substantially smaller that in intact narrative for the same number of retained clauses. For comparison we presented the theoretical prediction for the random list of words, which was shown to describe data well \cite{naim2020fundamental}. It is clear that there are more clauses recalled in intact narratives than words in lists of random words. Surprisingly, retrieval of scrambled stories is significantly {\it worse} than random lists, suggesting an active suppression of items in service of generating a coherent recall (participants were implicitly instructed to recall story). Finally, we show the mean number of clauses in a recall (green crosses), which is insensitive to the {\it content} of the clause and just measures the length of a participant's recall. For $R$ we report standard error in total recall length over the entire population of subjects; for $M$, we calculate error using bootstrap.}
	\label{fig:RvM}
\end{figure}

Having each narrative seen by roughly 200 participants, half of them doing recall and another half recognition, we were able to determine the average number of clauses retained in memory ($M$) and recalled clauses ($R$). As expected, both $M$ and $R$ grow with the length of the narrative presented, as measured by the number of clauses in the narrative and denoted by $L$ (see Fig. \ref{fig:RvM}A,B). Moreover, both $M$ and $R$ appear to grow linearly with $L$ for the range of narrative lengths we explored, and hence when we plot $R$ vs $M$, we also get an approximately linear relationship (see Fig. \ref{fig:RvM}C). This scaling behaviour is very different from what we observed with random lists of words with a characteristic square root scaling, i.e. unsurprisingly, recall of meaningful material is better than for random ones of the same size, even if we discount for better memorization. We should point out however that for the longest narrative in our set, some of the participants 'summarized' several clauses in the narrative into a single clause of recall (see also Discussion). Scoring of these instances is ambiguous, and the LLM often judged summarized clauses as all being recalled, resulting in the average number of recalled clauses ($R$) being substantially larger than the average number of clauses in recalls of this narrative ($C$; green crosses in \Cref{fig:RvM}C, see also \Cref{app:compression},  \Cref{fig:recall_compression}A). If we apply a more conservative scoring and use $C$ instead of $R$ as a measure of recall, linear scaling of recall will not persist when the longest narrative is included (\Cref{fig:RvM}C; \Cref{fig:recall_compression}B;  see also Discussion).  

\begin{figure}[ht!]
 \includegraphics[width = 1\textwidth]{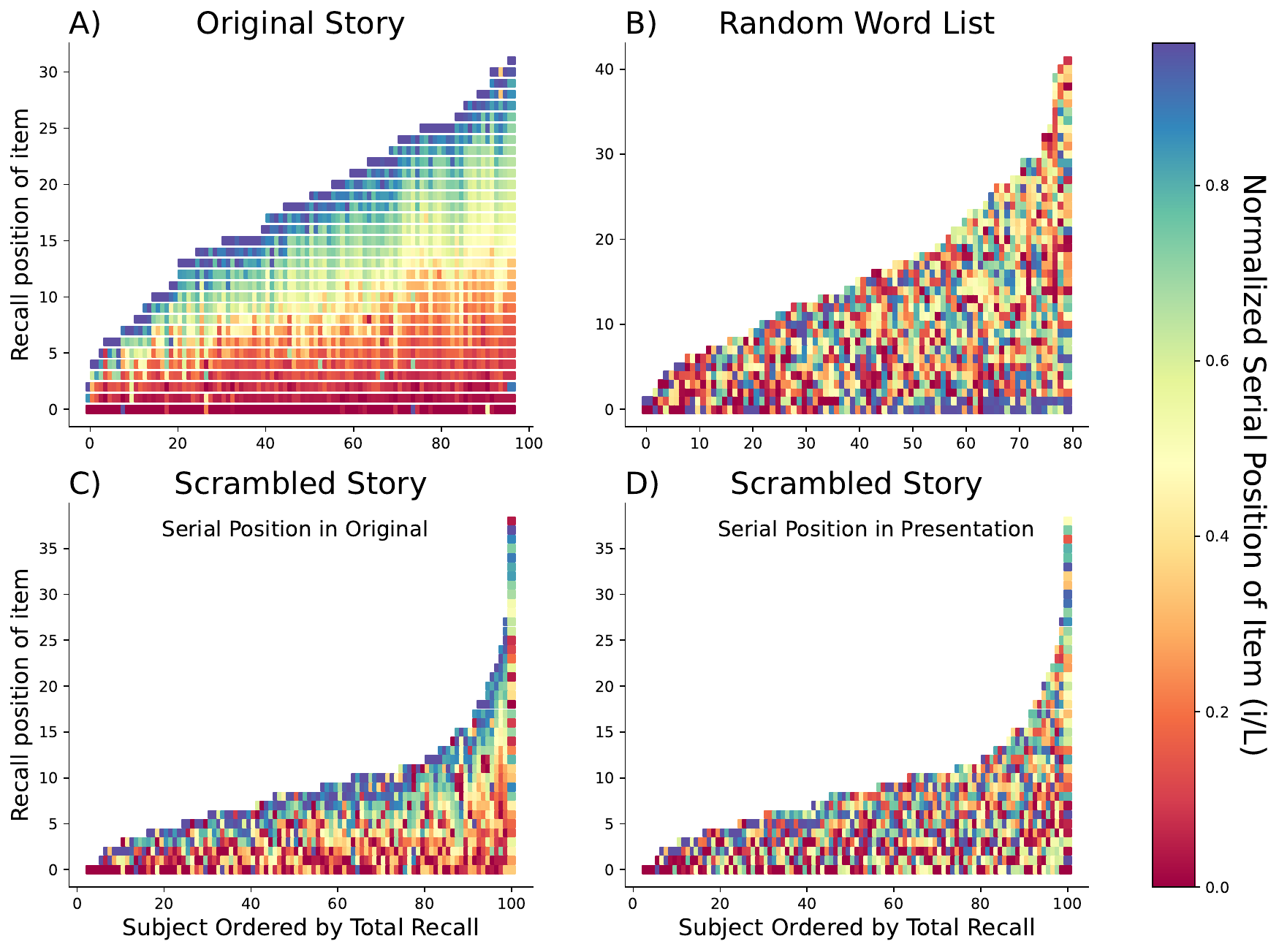}\
	\caption{ {\it Recall order}. Color-coded order of clauses or words for different conditions are shown in all panels. Recalled clauses or words are stacked together vertically (with the first recalled clause at the bottom of a column, and the last recalled clause at the top). The height of the column represents the total number of clauses or words recalled in a given trial. In panels A, B, and D, color code represents serial position of {\it presentation} of clauses or words, from early (red) to later (blue) in presentation position. Panel (C) is the only exception, in which the color code reflects the serial position of clauses in the {\it original} (intact) story. (A) shows that recall of coherent stories largely preserves presentation order. (B) recall of random word lists does not preserve presentation order. (C) As with random lists, the recall of a scrambled story does not preserve presentation order, but rather appear to reconstruct the original order of the story, as seen from the color gradients in panel (B).
 Apparently, random words and scrambled stories are recalled in random order considering their presentation order, but people perform some unscrambling of the scrambled stories as can be seen in (C) - there is tendency of recalled clauses being in the order of original unscrambled narrative. The participants construct a mental representation of the scrambled narrative which is evidently close to its original form. Recall consequently does not reflect input sequence, but rather the original sequence of the clauses.
}
	\label{fig:recall_order}
\end{figure}

One of the factors that apparently leads to better recall of narratives is the temporal ordering of recall. When people recall narratives, recall mostly proceeds in the forward direction (see Fig. \ref{fig:recall_order}A), probably reflecting the natural order of events in the narrative that cannot be inverted without affecting its coherence. This contrasts with the case of random lists, when recall proceeds in both directions with similar probability (see Fig. \ref{fig:recall_order}B), which, according to a model proposed in \citep{naim2020fundamental} results in the process entering a cycle preventing many words from being recalled. 

\subsection{Meaning and memory}
\begin{figure}[ht!]
 \includegraphics{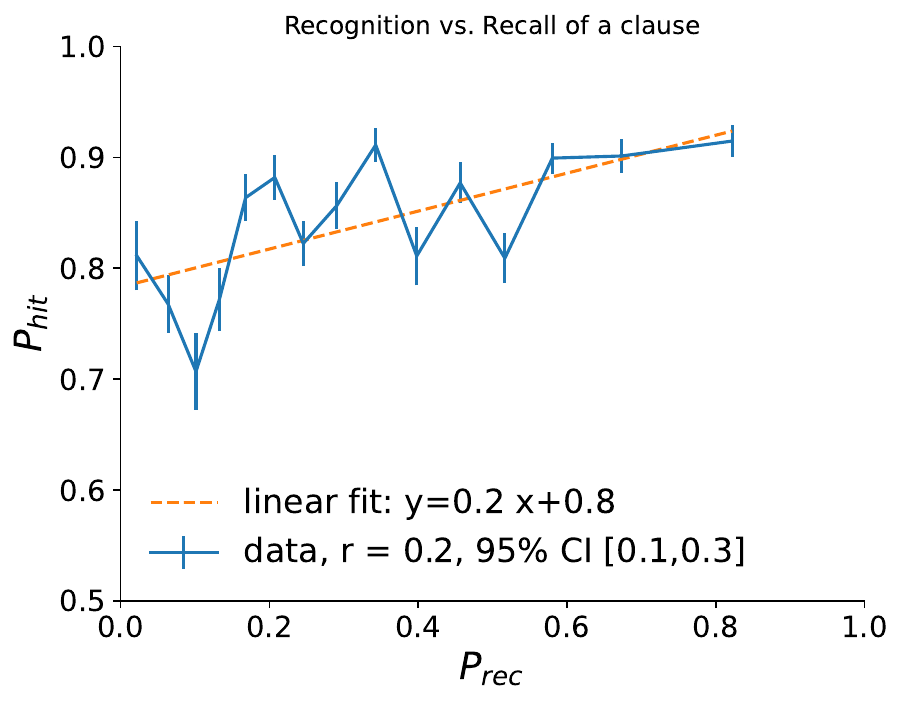}
 
\caption{ {\it Recognition vs recall performance across different clauses}. Clauses from all the narratives used in this study were divided evenly into 15 bins according to their $P_{rec}$, and the average $P_h$ for the clauses in each bin was computed and plotted against the center of the corresponding bin. Error bars show standard error within a bin. We show a linear fit (orange dashed) to the binned data (solid blue). The correlation coefficient $r$ is computed using the unbinned cloud of data points, with the $95\%$ confidence interval calculated using bootstrap with $3000$ samples.}
	\label{fig:p_hit_vs_P_rec}
\end{figure}

As we mentioned above, people's recall is strongly influenced by  narrative comprehension, such that clauses that are most important in communicating the summary of the narrative are the ones that are recalled by most of the participants. We found however that recognition is not so strongly affected by meaning. This can be observed by evenly dividing all the clauses used in our experiments into subsequent bins according to their recall probabilities, and calculating the average recognition performance for all the clauses in each bin. Surprisingly, there is very little increase of recognition with recall probability across the clauses, such that clauses with highest and lowest average $P_{rec}$ only differ in their $P_h$ by less than 0.15  (see Fig. \ref{fig:p_hit_vs_P_rec}; c.f. \citep{thorndyke1980critique,yekovich1981evaluation}). 

To further elucidate the role of meaning in memory, we repeated our experiments with another group of participants after randomly scrambling the order of clauses, thus making comprehension much more difficult if not impossible. We found that, unsurprisingly, recall of scrambled narratives is much poorer than the original ones (Fig. \ref{fig:RvM}B,C). Recognition performance for scrambled narratives however is practically the same (Fig. \ref{fig:RvM}A).  This result indicates that memory encoding of clauses is not significantly affected by the structure and meaning of the narrative.  Interestingly, the order in which people tend to recall clauses from a scrambled narrative corresponds much better to the order of these clauses in the original narrative than in the presented, scrambled one (see Fig. \ref{fig:recall_order}C,D), indicating that even in this situation people are trying to comprehend the meaning of the narrative rather than processing the input as a random list of unconnected clauses. This might explain why recall of scrambled narratives appears to be \textit{worse} than recall of random word lists of the same size.

How can recognition remain so strong even when recall is suppressed? We observe even for coherent narratives in \Cref{fig:p_hit_vs_P_rec} that recognition is generally quite high, even for $P_{rec}$ approaching zero. These observations appear to be consistent with the dual-process model of recognition \cite{yonelinas2002nature}, wherein familiarity judgments can be used to perform recognition even in the absence of explicit recollection (low recallability). 

\subsection{Semantic Similarity Predicts Recall}\label{sec:similarity}

One prominent feature seen in \Cref{fig:reliability}A is a wide range of recall probabilities for different clauses. In other words, while some clauses are recalled by most of the participants, other clauses are not. Such a wide distribution of $P_{rec}$'s across the clauses was observed in recalls of all the narratives and contrasts sharply with corresponding results for random lists of words where $P_{rec}$'s are rather uniform except for the ones in the beginning and the  end of the list (see \Cref{app:serial-position}). This wide distribution of $P_{rec}$'s is apparently due to the fact that not all clauses have similar importance for communicating the narrative. Indeed, if we select the clauses with high enough $P_{rec}$, we usually get a good summary of the narrative (see \Cref{sec:a_hprc} for examples). 

Can the recallability of a clause be predicted directly from a narrative? There is good reason to believe this is possible. Anecdotally, we know that highly skilled storytellers are often able to construct very memorable narratives, suggesting at least an intuitive understanding of the interplay between story structure and memory. Previous work has explored various methods to quantify this relationship. Most germane to our approach is \cite{Lee2022}, which introduces ``semantic centrality", a network measure of how semantically connected an event is to other events in a narrative. This was shown to correlate well with the probability to recall an event after watching a short video. Here, we introduce a simpler metric which directly computes the semantic similarity between a single clause and the entire narrative. This is motivated in part by the above-mentioned observation that the most highly recalled clauses, when read together in isolation, tend to make good summaries of the narrative (c.f. \Cref{sec:a_hprc}). This suggests that these clauses are more important for determining the meaning of the narrative as a whole, and should therefore score higher under any metric of semantic similarity to the entire narrative.

We test this hypothesis using LLM text-to-vector embeddings. The semantic similarity is taken to be the cosine similarity between the vector representation, or embedding, of a single clause and the vector embedding of the entire narrative (including the clause) (for details see \Cref{sec:sim_appendix}).

\begin{figure}[ht!]
 \includegraphics[width = \textwidth]{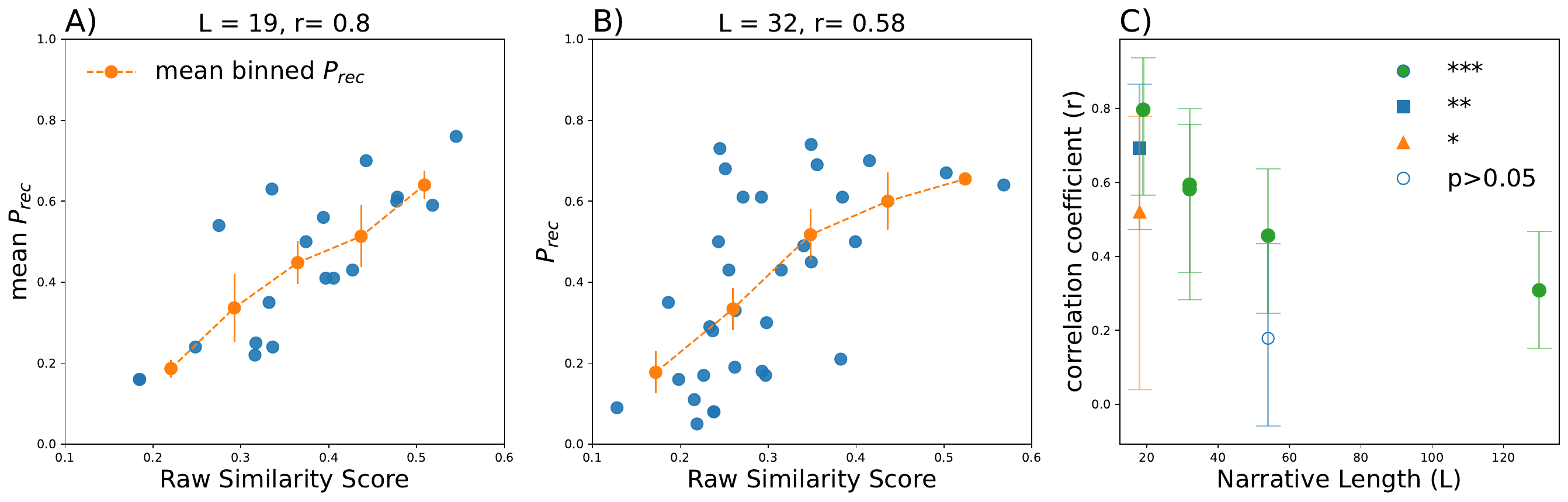}
 
\caption{ {\it Semantic similarity correlates with recall probability}. A) Scatter plot of recall probability $P_{rec}$ vs cosine similarity score (described in text) for each clause in a narrative with $L = 19$ clauses. Plotted in orange is the mean $P_{rec}$ and standard error per bin, for 5 bins, with the horizontal coordinate taken at the midpoint of the bin. The correlation coefficient ($r$-value) is 0.8 and statistically significant ($p<< 0.001$). B) Same as (A) with a story of length $L = 32$, and a statistically significant correlation of $r = 0.58$. Error bars are computed using standard error within in each bin. C) shows the correlation coefficient between $P_{rec}$ and similarity scores computed for each story, plotted here as a function of story length. The significance level is indicated in the legend, with green triangles indicating p value $p < 0.001$ (***), blue squares $p < 0.01$ (**), and orange triangles $p < 0.05$ (*), while empty circles indicate no statistically significant correlation ($p>0.05$). p-values are computed using two-sided Wald test. $95\%$ confidence intervals are computed using bootstrap with $1000$ samples, and indicated with capped error bars in the figure. Text embeddings for this figure were obtained using OpenAI's \ff{text-embedding-3-small} model. }
	\label{fig:similarity_scores}
\end{figure}

For each narrative of length $L$, we compute $L$ scores for each clause, which we subsequently compare to the recall probabilities. The main results of this analysis are presented in Fig.(\ref{fig:similarity_scores}), with a more thorough analysis saved for \Cref{sec:sim_appendix}. We find a very strong relationship between our semantic similarity score the recall probability for a majority of narratives. \Cref{fig:similarity_scores}A and B show two examples for a shorter and longer story, while \Cref{fig:similarity-score-all-narratives} shows the same data for all of the narratives we used in our study. \Cref{fig:similarity_scores}C shows that for individual narratives, the similarity score shows a statistically significant correlation with $P_{rec}$ in a majority of the narratives we tested. There is however a general trend of correlations getting lower with story length. We speculate that this is due to the general trend of subjects to resort to summarization with longer narratives, thus potentially making such a clause-level analysis of similarity an overall weaker predictor of recall performance. 

In \Cref{sec:sim_appendix}, we also compare different embedding models and their ability to predict recall via our similarity score. The embedding models that we study have varying success in predicting recall. This suggests a novel benchmark for text embeddings that directly reflects how humans process and ascribe significance or importance to parts of a text.


\section{Previous Work}
\label{sec:prev-work}

The experimental study of memory for narratives can be traced back to the highly influential descriptive work of \citet{bartlett}. This and followup work introduced the idea that the encoding of memory for narratives is a process of abstraction \citep{Gomulicki1956}, and that subsequent recall is in large part a generative process driven by a participant's prior knowledge and biases. This line of thought was formalized much later in theories of narrative structure involving schemas, scripts, frames, and story grammars \citep{alba1983memory,Rumelhart1975}. By now, there is ample support for an abstracting process for memory encoding, and the existence of schematic structures which guide recall \citep{baldassano2018representation}. 

A parallel line of research into narrative structure originated in the field of sociolinguistics by \citet{Labov1966}. In carrying out linguistic fieldwork to analyze spoken dialects of English, the authors found that personal narratives of emotionally charged events tented to elicit the best examples of local dialect. Furthermore, the resulting narratives, which were produced spontaneously and by non-expert storytellers, tended to be very regular in their structural properties. Of particular significance to us was the observation that stories were typically told in the order in which they experienced by the speaker, which Labov encapsulates in his ``egocentric principle" and ``no flashback'' constraint \cite{labov2013language}. This lends some support to the strong iconicity assumption \citep{zwaan1996processing}, which states that readers (or listeners) will interpret the order in which events are reported as reflecting the actual chronological order in which they occurred. 


However, serial order alone likely cannot explain the rich structure we observe in narrative recall. Indeed, temporal order provides just a single axis along which events in a narrative are organized. In addition to simple serial order, events or clauses can have causal relations (e.g. A causes B) \citep{Trabasso1985,Trabasso1985a, Lee2022}, inferential relations (e.g. A implies B), and superordinate or constituency relations (e.g. A consists of B) \cite{black1979episodes}. These relations conspire to give narratives hierarchical structure \citep{kintsch1998comprehension}. Story grammars provide one natural mechanism whereby the hierarchical structure arises, and served as early inspiration for studying structure dependence in narrative recall and recognition.  

In this direction, \citep{yekovich1981evaluation,thorndyke1980critique} performed recall and recognition experiments to test how encoding and processing depend on hierarchical structure. Their recognition experiments included only old, paraphrase, and false statements in the test phase. The false statements tested were appropriate (invoking the appropriate agents or actions), but inconsistent with or contradictory to the story. They observe that recognition is very weakly correlated with recall, and not significantly different from zero (Table 3 of \cite{yekovich1981evaluation}).Furthermore, whereas recall was argued to be sensitive to the structure of a narrative, recognition appeared to have no such sensitivity. 

In another experiment seeking to isolate the structure sensitivity of narrative processing, \citet{kintsch1977summarizing} tested processing of scrambled narratives, in which paragraphs of a text were presented to participants in random order. Participants were able to produce coherent summaries of such scrambled text, which were indistinguishable from summaries produced by participants reading the coherent text. It is possible that scrambling larger units (paragraphs in this case, versus clauses in our experiments) produced an overall more comprehensible text, making descrambling easier. Nevertheless, we still observe descrambling, but of lower quality (see \Cref{fig:recall_order}C). 

There have been many efforts to uncover what features of an event in a narrative make it more or less memorable than other events. \citet{Trabasso1985a} showed that the strongest predictor for recall was causal connectedness, in which more memorable clauses have more {\it causal} links to other clauses in the narrative. Some recent support for this observation was presented in \cite{Lee2022}, in which subjects performed immediate recall after watching short video clips. They showed that, in addition to causal centrality, a measure of {\it semantic centrality} was predictive of whether an event from the movie was later recalled. Crucially, semantic content was accessed using vector embeddings of text. 

Semantic vector embeddings have been used in psychological modeling since the introduction of latent semantic analysis (LSA) \cite{deerwester1990indexing}, a technique which obtains embeddings from a low-rank approximation of a matrix of word-document occurrences in a corpus. The psychological relevance of the resulting embeddings has been demonstrated in many settings, including semantic proximity effects in free recall \cite{howard1999contextual}, and similarity judgements \cite{bhatia2019distributed}, just to name a few. The hypothesis of distributional semantics, which states that the meaning of a word can be captured by its statistical properties in a text, has been pushed to extremes with modern deep learning, given the availability of huge datasets and computational power. The resulting semantic embeddings have displayed surprising properties, such as linear structure that allows analogies to be solved by vector algebra \cite{mikolov2013distributed}. These approaches have culminated recently in deep neural networks that can effectively embed entire chunks of text, opening up new avenues to study semantic structure of memory and free recall in narratives. For instance, \citet{heusser2021geometric} utilize text embeddings to study the trajectories of story recall in semantic space, showing how gross structure of a trajectory can remain stable across participants, with high-frequency deviations from this path indicating selective omission of details. 

Scaling laws for memory have been observed for random unstructured lists (pictures, words, etc) \citep{Lionel1973, murray1976standing, naim2020fundamental}. Two important takeaways from these works are the following: memory typically exhibits power-law scaling with list length, and the retrieval process appears to be universal. Surprisingly, there is very little work which considers the scaling of narrative memory with story length, let alone attempts to quantify it. The only work we are aware of is  \citep{Glenn1978}, which measured average recall as a function of story lengths of up to 83 ``informational units", which include clauses as well as noun and adjective phrases. The experimental design was motivated largely by questions about story grammars, and therefore the episodic structure of the narratives (in the story grammar sense developed by the authors) was kept constant while descriptive phrases and clauses (so called 'informational units') were added to increase the length of the story. Therefore, the added statements are arguably a kind of filler, not contributing any significant additional meaning or structure to the underlying narrative. This might account for the seemingly sublinear scaling of their mean recall with story length, compared to our linear results (\Cref{fig:RvM}). There are other significant differences, including the test population (24 second-grade schoolchildren), the stimulus input format (spoken narratives), and the recall format (spoken recall).

The choice of ``informational units" in \citep{Glenn1978} also differs from our choice of clauses, and is similar to segmentations that use propositional analysis \citep{Kintsch1978}, or pausal breaks \citep{Johnson1970}. An important feature of these different units is that they provide a more fine-grained segmentation of a narrative - a clause can consist of many propositions or pausal breaks, but not the other way around. However, these smaller units would not have a narrative function in the Labovian sense. There is also evidence from experiments on immediate verbatim recall that clauses are recalled holistically and largely intact \cite{jarvella1979immediate}. This suggests that beyond being a descriptive segmentation of text useful in the field of linguistics, clauses are psychologically processed as coherent units.  

Indeed, it is generally believed that humans process continuous naturalistic stimuli via discrete segments. \citet{radvansky2017event} review theory and experimental evidence in favor of ``events" as being the natural units of experience. In the Labovian classification, an event roughly corresponds to an episode, and so would constitute a more coarse-grained segmentation of a narrative \cite{bailey2017attentional}. \citet{sargent2013event} suggest that recall memory for a movie was correlated with an individuals ``segmentation ability", which measures how close their individual segmentation was to the average over a population. This natural tendency to segment stimuli, and its effects on recall memory, seems to give credence to our choice to measure memory for narratives in discrete units.

\section{Discussion}

In this contribution, we describe a new way to study human memory for meaningful narratives with the help of large language models. Together with using internet platforms for performing experiments, this technology enables a quantitative leap in the amount of data one can collect and analyze. In particular, we describe the prompts that we used to make LLMs generate new narratives of a particular type and size, score multiple human recalls of these narratives by identifying which clauses were recalled, and generating plausible lures for recognition experiments. Having a large amount of data is important for memory research because, as opposed to e.g. sensory processing that is believed to be largely uniform across people \citep{read2015psychophysics}, the way people remember and recall meaningful material is highly variable. Hence, only through statistical analysis can some general patterns be uncovered. In particular, we considered how recognition and recall performance scales up for narratives of increasing length. We found that approximately the same fraction of narrative clauses are recognized and recalled when narratives become longer, in the range of 20 to 130 clauses that we considered for this analysis. We expect that as narratives become longer, this trend will not persist because people will start summarizing larger and larger chunks of the narrative into single sentences in their recalls, as indeed happened for some of the participants recalling the longest narrative in our pool (see \Cref{app:compression}). When summarizing happens, deciding whether individual narrative clauses in the chunk were recalled or not is ambiguous. In the extreme case, a short summary of the entire story can be scored as having recalled nothing, since no particular clause was recalled, when in fact there was a non-trivial recall but at a higher level of abstraction or organization. This in itself does not mean that studying recall for clauses is incorrect or uninformative, but rather that it must have some regimes of validity. 

This discussion illustrates a fundamental limit in our analysis connected with the choice of linguistic clauses as the unit of recall and recognition. The basic issue at play is that the choice of segmentation imposes a natural scale into the analysis, which will inevitably interact with other scales in the task, such as the total length of the narrative (experiment duration), working memory capacity, or even participant engagement. An important observation which follows from the discussion in the previous paragraph, is that measuring recall at the scale of clauses will miss structure that emerges at a more coarse-grained scale, such as summaries. Capturing this structure might necessitate measuring larger units such as events, which is a standard choice in studies of memory for longer stimuli like movies or video clips, and which for text-based stimuli was shown to consist of many clauses (on the order of ~10 as found in \cite{bailey2017attentional}). It is also possible to resort to a more fine-grained analysis of text. Following previous work, we might consider a segmentation into propositions, which are abstract representations of the semantic content of a text. It has been argued, most persuasively by \citet{kintsch1998comprehension}, that the structure of memory is semantic, and representations are propositional. While a clause itself is a proposition, consisting of a predicate with arguments, the arguments themselves can, in certain cases, be composed of multiple propositions. So a single clause could represent a (likely shallow) hierarchy of propositions. We expect that if we repeat our analysis for such finer-grained units, we will see a similar phenomenon to the one we observe for clauses: fine-grained propositions will be replaced with more coarse-grained units as the story length increases, but this transition will happen earlier (i.e. for shorter stories) than what we observe for clauses. There is therefore a need to develop techniques which are sensitive to this multi-scale, hierarchical structure of recall, by automatically detecting higher-level units of meaning such as events and episodes, as well as possibly fine-grained units such as propositions.

We investigated the role of meaning in narrative memory by presenting the participants with the  same clauses as in the original narrative but in a scrambled order. These `narratives' are much more difficult, if not impossible, to make sense of, and indeed their recall was very poor. However, recognition of individual clauses was practically as good as in the original narrative. This surprising observation indicates that the encoding of clauses in memory is not very sensitive to the overall structure of the narrative, and only in the process of recall  does the meaning of the narrative play a major role. This finding is consistent with the observations of \citep{thorndyke1980critique,yekovich1981evaluation}. It also provides a striking confirmation of the generally held wisdom that while recall is structure sensitive, recognition need not be \citep{kintsch1998comprehension}. It is still possible, however, that these trends will change when longer narratives are considered. This will have to be investigated further. 

Another interesting observation concerns the fact that as the narrative becomes longer, the range of recall probabilities for different clauses remains very wide, e.g. there are always some clauses that are recalled by most of the participants and others that are almost never recalled. In contrast, the probability to recall words from a random list of words decreases with the length of the list, with the exception of the last few words (due to the recency effect) \citep{murdock1960immediate}.

An interesting theoretical question is to identify the factors that predict how well a given clause will be recalled in a given narrative. In this paper, we show that clauses which are semantically closest to the overall meaning of a narrative are best recalled. We accomplish this using semantic vector embeddings generated by LLMs. We show that the cosine similarity between  a clause embedding and the whole-narrative embedding is a strong predictor of recall probability. Therefore, the geometry of LLM embeddings reflect in some way how humans process and assign importance to different clauses. Being a relatively new technology, there is still much work to be done in understanding and interpreting the structure of embeddings produced by these models. Our work implies that the geometry of embeddings has direct relevance to human cognitive processing of meaningful text. 


We focused on first-person spoken stories. These were personal accounts of important events, shared naturally and informally. This way, they lacked the refinement of crafted stories, which may utilize tricks to improve memorability. It would be interesting to see how scaling of memory is affected by such expertly told or literary stories. Evidently there is an impact, considering that many stories in the oral tradition endure over very long timescales  \citep{rubin1995memory,nunn2018edge}. A striking example of this in a more controlled lab setting had participants trained to construct a narrative in which they embedded a random list of words \citep{Bower1969}. This work found that employing such a mnemonic improved recall to nearly perfect for up to 12 consecutively learned lists.

While practice and rehearsal are necessary for preserving narratives in oral tradition, our results suggest that narratives are intrinsically quite memorable. We find that memory performance for narratives encountered only once, as measured by the scaling relations in \Cref{fig:RvM}, are robustly superior to performance on unstructured lists.  

The research conducted for this report relied crucially on a set of LLM input ``prompts'', i.e. instructions, written in standard English, given to the LLM for carrying out various tasks. Roughly speaking, these appear as if they were instructions given to a human research assistant. Quite remarkably, the LLM completes the input string to provide a correct output without any additional training, a phenomenon  known as `in-context' learning \citep{brown2020language}. Since this phenomenon is still not fully understood, we had to resort to a good amount of trial-and-error and fine-tuning in designing the prompts used in our analysis. We provide all of the prompts used in our experiments in \cref{app:prompts_and+completions}. The specific model we utilize in most of the paper is OpenAI's GPT-4. However, while we believe the capabilities necessary to carry out our experiments are not limited to this model, it is an open question whether the prompts we use can be transferred to different models. 

In summary, using LLMs in conjunction with internet platforms for performing experiments is a powerful new tool that could bring significant new advances in understanding human memory.

\section{Methods}\label{methods}

With the aim of conducting a large scale study on memory for natural continuous material (personal narratives in this case) we required an automated procedure that would facilitate measuring human recall memory performance, since manual scoring of recalls is very labor-intensive and thus limits the ability to analyse large datasets. We were able to achieve this through the use of Large Language Models (LLMs) and we assessed the reliability of our pipeline by comparing it to human scoring performed by the authors. Our dataset was generated by conducting recall and recognition experiments online, recruiting participants through a crowd-sourcing platform. All segments of this study are detailed below.

\subsection*{Stimulus Set - Narrative pool}
\label{sec:narrative-stimuli}

Nearly all of the stimuli we use are generated by LLMs and are based on first-person oral narratives taken from socio-linguistic interviews \citep{Labov1966,labov2013language}\footnote{The narratives `boyscout' and `stein' are taken directly from these references. The rest are generated using templates.}. The oral narratives are segmented into clauses in these references, and these are used as templates for the LLM narrative generation. The LLM output is a narrative of equal length (in number of clauses), with very similar narrative-syntactic structure, but involving different subject matters. Two stories were generated from each template for lengths L = 18, 32, and 54. Two additional narratives were directly taken from Refs. \citep{Labov1966,labov2013language}, one with L = 19 (`boyscout' \ref{ex:boyscout}) which was analyzed in \cite{Labov1966}, and the other with L = 130 which was minimally edited to remove punctuation due to speech breaks, in order to increase readability (story of Gloria Stein in Ch. 7 of \citep{labov2013language}, which we do not reproduce below). We tested with four scrambled narratives at lengths $L = \{19,32,54,130\}$, by selecting a single story at each of these lengths, and randomly permuting the clauses. The resulting scrambled narratives then have an identical set of clauses to the original story, just presented in a random order.  Some basic statistical features of the stories (including word and character counts) are summarized in \Cref{sec:stimuli-stats}.

More details of the narrative generation by LLMs can be found in \cref{sec:narr_gen}, along with a sample narrative template in \ref{arg:schissel} and examples of generated outputs in \ref{comp:story_gen_1}. All but one of the narratives used as stimuli can be found in \Cref{sec:narratives}.

For the purpose of evaluating the reliability of recall scoring of LLMs and their similarity to human scoring, we generated and segmented a different narrative based on the `boyscout' story (\ref{ex:boyscout}, L=19). This stage began before the rollout of \ff{gpt-4} and for this reason we document the evaluation process separately in \Cref{sec:reliability}. The narrative generation step in this part produced variable length narratives in prose, which we had to subsequently segment using GPT-3. As a result, this narrative generation procedure did not keep the same number of clauses as the story it was based on (`boyscout').

\subsection*{Experimental Design}

Participants were recruited online through the `Prolific' platform (www.prolific.com) and experiments were conducted on participant's browser. For all experiments conducted in this study, ethics approval was obtained by the Institutional Review Board of the Weizmann Institute of Science, and each participant accepted an informed consent form before participation. Only candidates that indicated English as their first language were allowed to participate. No demographic information about the participants was collected. The trial was initiated by a button press. After a three second counter, a narrative was presented in the form of rolling text (marquee) in black font in the middle of a white screen. All narratives were animated in constant speed of 250 pixels per second. The average character size was approximately 22px, leading to a rate of approximately 12 characters per second.  
Once the marquee for the narrative has traversed outside the screen (all characters shown and disappeared), the testing phase was triggered automatically. This marquee style presentation was chosen because it allowed for comfortable reading while fixing the presentation duration for all participants and simultaneously preventing revisiting of already read material.


\paragraph{Free Recall Experiments}
In the free recall experiments, subjects were first shown the following written description: ``This is a {\bf recall task}. You will be shown a small narrative in the form of rolling text and then you will be prompted to write it down as you remember it. Try to include as many details as possible." The testing phase consisted of a textbox and a prompt to ``Please recall the story". Once participants finished typing their recall of the narrative, they submitted their response with a button press and the experiment was concluded. The number of participants we ran for each free recall experiment can be found in \Cref{sec:stimuli-stats}.

\paragraph{Recognition Experiments}

In the recognition experiments, subjects were first shown the following written description:``This is a {\bf recognition task}. You will be shown a small narrative in the form of rolling text and then you will be shown different clauses, one at a time and your task will be to choose whether it was shown in the text or not according to your memory." This initial ``learning phase" was identical to that for the free recall experiments. For the testing phase of the recognition experiment, 10 queries were sequentially presented. In each query, the participant was shown a single clause at random, either from the just presented narrative (old) or a lure (new). They were prompted to answer Yes or No to the question: ``Was the following clause presented in the story?". The number of participants we ran for each recognition experiment can be found in \Cref{sec:stimuli-stats}.

We did not observe any signatures of output interference \citep{criss2011output} (see \Cref{app:output_interference}) and therefore used all queries in the subsequent analysis. 
Lures were generated by asking the LLM to take a given narrative segmentation, and insert novel clauses between each existing clause. This ensures that the lures are distinct from the true clauses, but still fit within the overall context of the story. For instance, this avoids lures which might mention "dolphins" if the story is about boy scouts. The prompt used to generate lures and an example completion by \ff{gpt-4} are given in \Cref{sec:lures}.

The recall and recognition experiments appeared on Prolific as separate experiments, and we generally got no overlap in participants (who were identified only by a unique hash string assigned by Prolific) for recall and recognition of the same story. However, for the $L = 32$ stories, there was non-negligible overlap in participants. Excluding the participants who performed the recognition experiments after the recall experiments for the same story (or vice versa), we found marginal changes in the summary statistics (e.g. $R$ and $M$) reported in the manuscript. Therefore, for the data reported in the figures, we use data from every participant in every experiment.

\subsection*{Analysis}
Analysis was conducted through custom Python scripts. 
For recognition memory, in order to estimate total encoded memory $M$ from \Cref{recog}, we used population and test trial averaged hit rate (true positive probability $P_{h}$) and false alarm rate (false positive probability $P_{f}$). Standard error was computed using statistical bootstrap \citep{efron1994introduction}.

Recall scoring was done using the OpenAI model \ff{gpt-4-0613} (a GPT-4 model which receives no updates) based on the clause segmentation of the narratives. For each participant's recall, \ff{gpt-4-0613} was instructed to loop through each clause of the given narrative (as presented) and examine whether the information that this clause provided was present in some form in the participant's recall and the corresponding passage. The numbers of all clauses evaluated as being recalled were given at the end of the output in the form of a list. The full prompt we use for scoring recalls is given in \ref{prompt:scoring}, which takes three arguments: the narrative stimulus in prose (e.g. \ref{arg:boyscout}), the numbered clause segmentation of the narrative stimulus (e.g. \ref{arg:boyscout_seg}), and the participant's recall (e.g. \ref{arg:boyscout_recall}). A sample completion is provided in \ref{comp:boyscout_score}.

Separately, to evaluate the similarity of recall scoring between humans and LLMs, 3 authors performed manual scoring of 30 recalls of \ref{ex:panic} using the same procedure, evaluating whether each clause was present in the recall or not. The data was collected under the free recall experimental protocol described above.  

For vector embeddings of text in \Cref{sec:similarity}, we used  OpenAI's embedding model \ff{text-embedding-3-small}. Comparisons with other embedding models are given in \Cref{sec:sim_appendix}.

\subsection*{Random List of Nouns Experiment}

In addition to the experiments which used narrative stimuli, we also performed an experiment with a list of 32 nouns (see \Cref{sec:list-of-nouns}) that were randomly selected from the pool of nouns used in \citep{naim2020fundamental}. The experimental protocol was exactly the same as in that work with presentation speed 1.5 sec per word. 105 participants were recruited using the Prolific online platform, with each participant accepting informed consent prior to the beginning of a trial.

\subsection*{Data And Code Availability}
Experimental data as well as code for the online experiments and the analysis presented above can be found in \url{https://github.com/mkatkov/llm-narrative-analysis}.

\subsection*{Author Contributions}
All authors designed the experiments. AG ran experiments. AG and TC developed the LLM pipeline for experimental design and data analysis. All authors wrote the manuscript.

\section*{Acknowledgements}
AG is supported by the Martin A. and Helen Chooljian Member in Biology Fund and the Charles L. Brown Member in Biology Fund. TC acknowledges the support of the Eric and Wendy Schmidt Membership in Biology, the Simons Foundation, and the Starr Foundation Member Fund in Biology at the Institute for Advanced Study, where this work was completed. MT is supported by the Foundation Adelis and the Simons Foundation. MK is supported in part by a grant from Fran Morris Rosman and Richard Rosman. We thank Omri Barak, Danqi Chen, Michael Douglas, Ariel Goldstein, and Weishun Zhong for helpful conversations. We thank Stefanos Gkouveris for  helpful comments on Javascript development. MT thanks Yadin Dudai for bringing the studies of spoken narratives by William Labov to his attention.

\bibliographystyle{apalike}
\bibliography{Memory} 
\clearpage

\appendix
\addcontentsline{toc}{section}{Appendix} 
\renewcommand{\partname}{\fontsize{28}{18} Appendix}
\part{} 
\parttoc 

\section{Methods}
Here we describe our experiments and data analysis methods in more detail.

\subsection{Stimuli Statistics}\label{sec:stimuli-stats}

\begin{table}[ht!]
	\caption{Coherent story stimuli: the first column lists the story ID in our dataset, and the following columns show: story length in clauses ($L$), approximate number of words ($L_{{\rm words}}$), string length of full narrative (i.e. number of characters $L_{{\rm char}}$), and an estimate of the total duration (in seconds) of learning phase of the experiment $T$. The final two columns show the number of participants we ran for the free recall ($N_{R}$) and the recognition ($N_{M}$) experiments with the corresponding story.}
	\centering
\begin{tabular}{c|cccccc}
   {\rm story} &  $L$ &  $L_{{\rm words}}$   &   $L_{{\rm char}}$  & $T ({\rm sec})$   & $N_{R}$ & $N_{M}$      \\ \hline
{\rm schissel-v1} & 18 & 147 & 803 & 67 & 56 & 100\\

{\rm schissel-v2} &18  & 149 & 775& 65 & 100 & 100 \\
{\rm boyscout} &19  & 142 & 711& 59 & 100 & 99 \\
{\rm triplett-v1} &32  & 254 & 1329& 111 & 100 & 100 \\
{\rm triplett-v2} &32  & 256 & 1291& 108& 101 & 100 \\
{\rm hester-v1} &54  & 478 & 2414& 201 & 98 & 99 \\
{\rm hester-v2} &54  & 617 & 3226& 269 & 100 & 99 \\
{\rm stein} &130  & 1469 & 7323& 610 & 99 & 99 \\
{\rm panic} & 56 & 374 & 1890 & 158 & 100 & 
\end{tabular}
\label{table:storystats_coherent}
\end{table}

\begin{table}[ht!]
	\caption{Scrambled story stimuli: the first column lists the name of the story which was chosen for scrambling at the level of clauses. The next columns show: story length in clauses ($L$), approximate number of words ($L_{{\rm words}}$), string length of full narrative (i.e. number of characters $L_{{\rm char}}$), and an estimate for the total duration (in seconds) of learning phase of the experiment $T$. The final two columns show the number of participants we ran for the free recall ($N_{R}$) and the recognition ($N_{M}$) experiments with corresponding story. }
	\centering
\begin{tabular}{c|cccccc}
   {\rm story} &  $L$ &  $L_{{\rm words}}$   &   $L_{{\rm char}}$  & $T ({\rm sec})$  & $N_{R}$ & $N_{M}$        \\ \hline
{\rm boyscout} &19  & 142 & 715& 60 & 100 & 100 \\
{\rm triplett-v1} &32  & 254 & 1333& 111  & 100 & 100\\
{\rm hester-v2} &54  & 618 & 3227& 269 & 101 & 99 \\
{\rm stein} &130  & 1425 & 7176& 598 & 100 & 99 \\
\end{tabular}
\label{table:storystats_scrambled}
\end{table}

Our main unit of measurement for the length of a narrative is the number of clauses. For completeness, we present here further statistics of the stories used in our experiments. We compute an estimate of the number of words by removing all punctuation and counting the number of text units separated by spaces. The total number of characters is computed using string length in Python, which includes all punctuation as well as spaces. The character count is used to estimate the total duration of the learning phase, the part of the experiments in which subjects read the story in a scrolling marquee which moves characters across the screen at a rate of approximately 12 characters per second. 

Note that the the scrambled version of the stein story has a different number of words than the coherent version. This is due to a processing step we applied before scrambling, wherein we cleared up some typographical quirks of the original transcription of stein. For instance, transcriptions a stutter such as "I - I guess"  were replaced with "I guess". This processing left the number of clauses invariant, and left the meaning of each clause unchanged. As such, it does not affect our analysis in the manuscript.

Finally, we included the details of the ``panic" story which was only used for reliability scoring in our manuscript (see Fig.\Cref{fig:reliability} and \Cref{sec:reliability})

\subsection{Prompts and Completions}
\label{app:prompts_and+completions}

A significant part of the successful use of LLMs is the effective design of inputs or ``prompts'', which amount to instructions, written in standard English, for carrying out a particular task. 

More precisely, the input to a LLM is a string, which we interchangeably refer to as the prompt or the context. The output of the LLM is also a string. This is also referred to as a ``completion", since LLMs are trained to complete text fragments given to them as inputs.

Despite the accumulation of wisdom on prompt engineering, we still resorted to a good amount of trial-and-error and fine-tuning in designing the prompts used in our analysis. Since these prompts are tantamount to algorithms written for LLMs, we present them below in pseudocode boxes. In the following sections, we exhaustively detail the prompts used, along with examples of LLM outputs (i.e. completions). The specific model we utilize in most of the paper is OpenAI's GPT-4. For the data analysis, in the interest of reproducibility, we opted to use the deprecated model \ff{gpt-4-0613} which does not receive updates. In \cref{sec:drift}, we show how scoring using the latest model \ff{gpt-4} can change with time as a result of OpenAI's regular model updating. 

\subsubsection{Narrative Generation}\label{sec:narr_gen}

We started with a template narrative, selected from the collection of oral narratives in \citep{labov2013language}, and instructed the LLM to produce variations of the story which changed the surface form, but kept the overall structure (e.g. number of clauses). For these stories, we used the segmentation given by \citet{labov2013language}, wherein the narratives are segmented word-for-word into linguistic clauses.  Here is a sample of the prompt, with arguments shown presented in a box enclosed in brackets, like  $\{ \ff{this}\}$.
\begin{prompt}[prompt:narr_gen]{Narrative Generation}
\code{This is a true personal narrative about a single event in someone's life. It has exactly $\{ \ff{N} \}$ clauses:}\\
\{ \ff{template narrative} \}\\

\code{Generate a new personal narrative that is unique and about something completely different. Try to keep the overall narrative structure of the personal narrative above, but change as much of the subject matter and action as possible. Do not just use the narrative and replace key persons, places and things. Make it completely new. This new narrative must also contain exactly $\{ \ff{N} \}$ clauses.}
\end{prompt}

As an example, we used the story of Jacob Schissel from \citep{labov2013language} as a template narrative which has $\ff{N} = 18$:

\begin{argument}[arg:schissel]{Example \ff{template narrative}}
\code{1. My brother put a knife in my head. \\
2. This was just a few days after my father had died \\
3. and we were sitting shiva. \\
4. And the reason the fight started,\\ 
5. he saw a rat out in the yard or \\
6.  – this was out in Coney Island – \\
7. and he started talk about it. \\
8. And my mother had just sat down to have a cup of coffee \\
9. and I told him to cut it out. \\
10.  'Course kids, y'know, he don't hafta listen to me. \\
11. So that's when I grabbed his arm \\
12. and twisted it up behind him. \\
13. When I let go his arm, there was a knife on the table, \\
14. he just picked it up \\
15. and he let me have it. \\
16. And . . .; I started bleeding – like a pig. \\
17. And naturally first thing to do, run to the doctor, \\
18. and the doctor just says, “Just about this much more,” he says, “and you'd a been dead.
}
\end{argument}

The output of the LLM upon processing this prompt was a story which clearly bore a structural relation to the original, but contained different agents involved in a completely different scenario. We provide an example completion below obtained using $\ff{gpt-4}$ with temperature $T = 0.6$. This example was used in experiments, and also appears in \ref{ex:schissel-v1-pool} next to another generated narrative ( \ref{ex:schissel-v2-lake}).

\begin{completion}[comp:story_gen_1]{from \ref{prompt:narr_gen}, \ref{arg:schissel}}

\code{1. My best friend pushed me into the pool. \\
2. This was right after I had broken up with my boyfriend \\
3. and I was feeling really down. \\
4. The reason she did it, \\
5. she thought I needed a distraction or \\
6. - this was at her house in the suburbs - \\
7. and she wanted to cheer me up. \\
8. I had just sat down with a plate of snacks \\
9. and I told her I was fine. \\
10. 'Course friends, y'know, they don't always believe you. \\
11. So that's when she sneaked up behind me \\
12. and gave me a little shove. \\
13. When I surfaced from the water, there was a float nearby, \\
14. she just grabbed it \\
15. and tossed it to me. \\
16. And . . .; I started laughing - uncontrollably. \\
17. And naturally the first thing to do, hug my friend, \\
18. and I just said, “Just a little more,” I said, “and I would've been drowning in my sorrows.
}

\end{completion}

The resulting narratives generated from this procedure vary in their coherence. For our experiments, we chose the most coherent generated narratives.

\subsubsection{Lure Generation}
\label{sec:lures}


We utilized \ff{gpt-4} at $T = 0.3$ with the following prompt:

\begin{prompt}[prompt:lure]{ Lure Generation}
\code{$\{$\ff{segmentation}$\}$\\
The items above all fit together to tell a story. Add more items of roughly the same length, numbered 1.5, 2.5, and so on, interleaving the existing items, elaborating on the story, and without repetition. These new items should introduce completely new plot elements, but still make sense in the context of the rest of the story. Add as many items as possible.
}
\end{prompt} 

The following is an example completion using the template narrative segmentation (\ref{arg:schissel}). For purposes of presentation, we present the lures along with the original clauses from the narrative. The completion does {\it not} include the original clauses, which appear in gray, and only consists of the text shown in black.

\begin{completion}[comp:lure]{from \ref{prompt:lure}}
\code{\\
\gray{1. My brother put a knife in my head.} \\
1.5. It was a tense time for our family, with everyone on edge.\\
\gray{2. This was just a few days after my father had died}\\
2.5. We were all grieving in our own ways, trying to cope with the loss.\\
\gray{3. and we were sitting shiva.}\\
3.5. Our house was filled with relatives and friends, offering their condolences.\\
\gray{4. And the reason the fight started,}\\
4.5. My brother had always been a bit of a troublemaker, but I never thought he'd go this far.\\
\gray{5. he saw a rat out in the yard or}\\
5.5. He was always fascinated by the creatures that roamed the streets and alleys.\\
\gray{6.  – this was out in Coney Island – }\\
6.5. The sun was setting, casting an eerie glow on the scene.\\
\gray{7. and he started talk about it. }\\
7.5. He started describing the rat in gruesome detail, trying to get a rise out of us.\\
\gray{8. And my mother had just sat down to have a cup of coffee}\\
8.5. She looked so tired and worn out, I just wanted her to have a moment of peace.\\
\gray{9. and I told him to cut it out. }\\
9.5. I tried to keep my voice calm, but I could feel my anger rising.\\
\gray{10.  'Course kids, y'know, he don't hafta listen to me. }\\
10.5. He just smirked at me, clearly enjoying the attention.\\
\gray{11. So that's when I grabbed his arm }\\
11.5. I didn't want to hurt him, just get him to stop his nonsense.\\
\gray{12. and twisted it up behind him.}\\
12.5. He let out a yelp of pain, but I didn't think I'd done any real damage.\\
\gray{13. When I let go his arm, there was a knife on the table, }\\
13.5. I didn't even see him reach for it, he was so quick.\\
\gray{14. he just picked it up }\\
14.5. I barely had time to react before the blade was coming towards me.\\
\gray{15. and he let me have it. }\\
15.5. The pain was sharp and intense, like nothing I'd ever felt before.\\
\gray{16. And . . .; I started bleeding – like a pig. }\\
16.5. I could feel the warmth of my blood soaking through my clothes.\\
\gray{17. And naturally first thing to do, run to the doctor,}\\
17.5. I stumbled out of the house, desperate for help.\\
\gray{18. and the doctor just says, “Just about this much more,” he says, “and you'd a been dead.}\\
18.5. I couldn't help but think how close I'd come to joining my father.
}
\end{completion}

\subsubsection{Recall Scoring}\label{app:recall_scoring}

In order to score recall, we utilize zero-shot prompting with OpenAI's deprecated model \ff{gpt-4-0613}. We chose to use this particular model for the benefit of reproducibility, since we have found that results from the following analysis can drift in time as OpenAI updates its model (see \Cref{sec:drift}).

The recall scoring \ref{prompt:scoring}, shown below, was a block of text, represented as a python string, which took as arguments the original narrative, the numbered segmentation of the narrative, and a single participant's recall of the narrative. The prompt was constructed so that it would identify which clauses from the narrative are present in the recall. A consequence of this scoring procedure is that a single clause in the recall can encode multiple clauses from the original narrative. \ff{gpt-4} appeared to have no trouble identifying such compressed clauses in the recall, and frequently would identify multiple original clauses within a single clause in the participant's recall. An example of this can be seen in \ref{comp:boyscout_score}, where the same passage in the recall  (``even the scoutmaster looked on") is counted as recalling both clauses 15 and 16 from the original narrative (\ref{arg:boyscout}).

Below, we provide the scoring prompt, as well as some example arguments, and the resulting LLM completion. 

\begin{prompt}[prompt:scoring]{Recall Scoring}

\code{This is the original text:}\\
\{\ff{narrative}\}
\\

\code{It can be broken down into the following independent pieces of information:}\\
\{\ff{segmentation}\}\\

\code{Here is an alternative version of the original text where some of the above pieces of information may be missing:}\\
\{\ff{recall}\}\\

\code{For each of the numbered information pieces of the list above, evaluate whether the information of each piece is given in the alternative version of the story, stating the number and showing the corresponding passage from the alternative story it is given in. After, write all the numbers of the pieces that are given in the alternative version of the story in a set of brackets at the end of the response.}

\end{prompt}

\begin{argument}[arg:boyscout]{Example \ff{narrative}}
\code{Yeah, I was in the boy scouts at the time. And we was doing the 50-yard dash, racing, but we was at the pier, marked off and so we was doing the 50-yard dash. There was about 8 or 9 of us, you know, going down, coming back. And going down the third time, I caught cramps and I started yelling "Help!", but the fellows didn't believe me, you know. They thought I was just trying to catch up, because I was going on or slowing down. So all of them kept going. They leave me. And so I started going down. Scoutmaster was up there. He was watching me. But he didn't pay me no attention either. And for no reason at all there was another guy, who had just walked up that minute... He just jumped over and grabbed me.}
\end{argument}

\begin{argument}[arg:boyscout_seg]{Example \ff{segmentation}}
\code{1. Yeah, I was in the boy scouts at the time. \\
2. And we was doing the 50-yard dash, \\
3. racing, \\
4. but we was at the pier, marked off, \\
5. and so we was doing the 50-yard dash. \\
6. There was about 8 or 9 of us, you know, going down, coming back. \\
7. And, going down the third time, I caught cramps\\
8. and I started yelling "Help!", \\
9. but the fellows didn't believe me, you know. \\
10. They thought I was just trying to catch up, because I was going on or slowing down. \\
11. So all of them kept going. \\
12. They leave me. \\
13. And so I started going down. \\
14. Scoutmaster was up there. \\
15. He was watching me. \\
16. But he didn't pay me no attention either. \\
17. And for no reason at all there was another guy, who had just walked up that minute...\\
18. He just jumped over \\
19. and grabbed me.}
\end{argument}


\begin{argument}[arg:boyscout_recall]{Example \ff{recall}}

\code{A boy was with the scoutmaster and he was going up and down and on the third time of going up the pier he had cramps and shouted 'help' but no-one believed him, even the scoutmaster looked on by until this guy came out of nowhere and grabbed the boy and saved him.
}
\end{argument}

\begin{completion}[comp:boyscout_score]{from \ref{prompt:scoring}}

\code{1. Not given  \\ 
2. Not given  \\ 
3. Not given  \\ 
4. Not given  \\ 
5. Not given  \\ 
6. Not given  \\ 
7. Given - "on the third time of going up the pier he had cramps"  \\ 
8. Given - "and shouted 'help'"  \\ 
9. Given - "but no-one believed him"  \\ 
10. Not given  \\ 
11. Not given  \\ 
12. Not given  \\ 
13. Not given  \\ 
14. Given - "was with the scoutmaster"  \\ 
15. Given - "even the scoutmaster looked on"  \\ 
16. Given - "even the scoutmaster looked on"  \\ 
17. Given - "until this guy came out of nowhere"  \\ 
18. Not given  \\ 
19. Given - "and grabbed the boy and saved him"  \\ 
  \\ 
(7, 8, 9, 14, 15, 16, 17, 19)  
}
\end{completion}

We note here that the completion does not always appear as in this example. Some minor deviations might occur (e.g. parentheses replaced with brackets), or the format of the answers change (e.g. instead of ``Not Given", some completions give "There is no corresponding passage for ....").

\subsubsection{Ordered Recall Scoring}\label{sec:ordered_scoring}

In order to extract the order in which the original clauses appeared in a participant's recall, we added an extra step to the recall scoring discussed above. For a given narrative, segmentation, and recall, we concatenated \ref{prompt:scoring} with its \ref{comp:boyscout_score}, and appended to this the following additional prompt:

\begin{prompt}[prompt:order]{Ordered Recall Scoring}
\code{Now repeat the alternative version of the narrative with the number of the independent piece of information inserted next to the location in which it appears in the alternative version. Then, list the numbers separately in the order in which they appear in the alternative story immediately above. The final list of numbers should be enclosed in parentheses.}
\end{prompt}

Therefore, the total input or prompt for ordered recall scoring would be \ref{prompt:scoring} + \ref{comp:boyscout_score} + \ref{prompt:order}. The output for this example was the following:

\begin{completion}[comp:boyscout_order]{from \ref{prompt:scoring} +  \ref{comp:boyscout_score} + \ref{prompt:order}}

\code{A boy (14) was with the scoutmaster and he was going up and down and on the third time of going up the pier he had cramps (7) and shouted 'help' (8) but no-one believed him (9), even the scoutmaster looked on (15, 16) by until this guy came out of nowhere (17) and grabbed the boy and saved him (19).\\
\\
(14, 7, 8, 9, 15, 16, 17, 19)
}

\end{completion}

\subsection{Narratives used in the study}
\label{sec:narratives} The narratives that were generated by GPT-4 for this study used an existing segmentation into clauses. The original segmentation was taken from \citep{Labov1966, labov2013language}. These segmentations were provided to the LLM to generate new variants, as detailed in \cref{sec:narr_gen}. For each narrative template, we generated about 10 variants, and selected two for use in our experiments, based on our judgment of their coherence. With the exception of the longest narrative ($L = 130$), we present below all narratives that were presented to participants in our experiments. In the examples, the narratives are labeled by a prefix reflecting the source story, with the exception of ``panic attack" which was created by GPT-3. The scrambled narratives are based on intact narrative (without ``-scrambled'' in the narrative name) where all clauses were shuffled. The ``panic'' story was generated based on the ``boyscout'' story from \citep{Labov1966}, \ref{ex:boyscout} below. The story was generated using \ref{ex:panic_seg_prompt}.

\newpage
\ref{ex:schissel-v1-pool} and \ref{ex:schissel-v2-lake} below were generated using \ref{prompt:narr_gen}, with the Jacob Schissel story \ref{arg:schissel} from \citep{labov2013language} as a template.
\begin{multicols}{2}

\begin{story_example}[ex:schissel-v1-pool]{Narrative schissel-v1-pool }
\begin{enumerate}[leftmargin = *,itemsep=0mm]
\item My best friend pushed me into the pool.
\item This happened during my cousin's wedding reception,
\item where everyone was dressed to impress.
\item And the reason it happened,
\item she spotted a bee hovering near my face,
\item - this was at a fancy hotel garden -
\item and she tried to save me from being stung.
\item And my aunt had just handed me a glass of champagne,
\item and I warned her to be careful.
\item 'Course friends, y'know, they don't always think things through.
\item So that's when she gave me a little shove,
\item and I tumbled into the water.
\item When I resurfaced, gasping for air,
\item she just started laughing,
\item and she apologized profusely.
\item And... my beautiful dress was ruined,
\item and naturally, the first thing to do was to get out and dry off,
\item and my cousin just says, "Just about a few inches more," she says, "and you'd have landed on the cake."
\end{enumerate}\end{story_example}

\columnbreak

\begin{story_example}[ex:schissel-v2-lake]{Narrative schissel-v2-lake }
\begin{enumerate}[leftmargin = *,itemsep=0mm]
\item My best friend pushed me into the lake. 
\item This happened during our annual summer camping trip 
\item and we were enjoying a peaceful afternoon. 
\item The reason it all began, 
\item she spotted a frog near the water's edge or 
\item - this was at our favorite spot in the woods - 
\item and she dared me to catch it. 
\item I had just sat down to eat my sandwich 
\item and I told her I wasn't interested. 
\item 'Course friends, y'know, they don't always take no for an answer. 
\item So that's when she snuck up behind me 
\item and gave me a little nudge. 
\item When I emerged from the water, there was a frog on the shore, 
\item she just scooped it up 
\item and handed it to me. 
\item And . . .; I couldn't help but laugh – uncontrollably. 
\item And naturally, the first thing to do was to get revenge, 
\item and I said, "Just you wait," I said, "you'll get yours soon enough."
\end{enumerate}\end{story_example}

\end{multicols}

\newpage

\ref{ex:triplett-v1-rookie} and \ref{ex:triplett-v2-catlady} below were generated using \ref{prompt:narr_gen}, with the Charles Triplett story from \citep{labov2013language} as a template.
\begin{multicols}{2}
\begin{story_example}[ex:triplett-v1-rookie]{Narrative triplett-v1-rookie }
\begin{enumerate}[leftmargin=*,itemsep=0mm]
\item Back in the days when I was a rookie in the police force, the Chief was a veteran.
\item He had caught a notorious criminal –
\item or had been awarded for bravery.
\item But he had – a young daughter.
\item And in those days I was a fitness enthusiast.
\item And it seemed she was trying to impress me.
\item I never observed it.
\item Truth is I wasn't very fond of her because she was –
\item She was a charming girl until she opened her mouth.
\item She was a chatterbox.
\item Goodness, she could talk.
\item Then she left a message one day saying she was going to run away because he was always scolding her about me.
\item He arrived at my apartment.
\item Big intimidating look on his face.
\item I managed to calm him down.
\item and proposed “Well we'll search for her
\item and if we can't locate her well you can – do whatever you think is right.”
\item I was strategizing.
\item So he accepted my offer.
\item And we went to where they found her scarf – near a park.
\item and we traced down a little more
\item and we couldn't locate her.
\item And returned
\item – it was a police station –
\item she was sitting on a chair with a book in her hands.
\item She hadn't run away.
\item But – nevertheless – that resolved the issue for that day.
\item But that evening the deputy, Frank Mitchell, said “You better transfer and leave because that old man never forgets anything once he gets it into his mind.”
\item And I did.
\item I transferred.
\item and left.
\item That was the end of my rookie year.
\end{enumerate}\end{story_example}

\columnbreak

\begin{story_example}[ex:triplett-v2-catlady]{Narrative triplett-v2-catlady }
\begin{enumerate}[leftmargin=*,itemsep=0mm]
\item In the neighborhood I grew up in, Mrs. Baker was the resident cat lady.
\item She had adopted at least twenty –
\item or had given them homes.
\item But she had – a young grandson
\item and back then I was quite the tomboy.
\item And apparently, he was trying to befriend me.
\item I never caught on.
\item Truth is I didn't take to him much because he had –
\item He was a charming lad until you saw his manners.
\item He had poor manners.
\item Goodness gracious he had poor manners.
\item Then he left a note one day saying he was going to run away because she was always nagging about me.
\item She came to my treehouse.
\item Big old tabby cat in tow.
\item I talked her down
\item and said “Well we'll go find him
\item and if we can't locate him, you can – go ahead and call the police if you need to.”
\item I was strategizing.
\item So she took my advice.
\item And we went to where they found his baseball cap – near the old mill
\item and we trailed a bit more
\item and we didn't find a trace.
\item And returned
\item – it was a treehouse club –
\item he was sitting on a branch with a comic book in his hand.
\item He hadn't run away.
\item But – however – that resolved it for that day.
\item But that evening the local sheriff, Officer Dawson said “You better lay low and stay out of sight because that woman never forgets anything once she gets it into her mind.”
\item And I did.
\item I laid low
\item and stayed out of sight.
\item That was the second time.
\end{enumerate}\end{story_example}
\end{multicols}


\ref{ex:hester-v1-park} and \ref{ex:hester-v2-church} below were generated using \ref{prompt:narr_gen}, with the Adolphus Hester story from \citep{labov2013language} as a template.

\begin{story_example}[ex:hester-v1-park]{Narrative hester-v1-park }
\begin{enumerate}[itemsep=0mm]
\item  Let me share a story from when I was twenty-one years old
\item  I was living in a small apartment near the city center
\item  I had just started my first job
\item  And uh I was working as a graphic designer
\item  My father, who was recovering from a surgery, hadn't been out in – in uh – well I'd say about eight months
\item  He had to use a wheelchair to get around
\item  The only person there with him during the day was a kind neighbor, a lady in her sixties
\item  My younger sister was away at college –
\item  so it was just my dad and the neighbor lady at home during the day
\item  One day, I got this strong feeling while I was at work
\item  It told me: "If you go to the nearby park during your lunch break and pray for your father's recovery, he'll start to improve
\item  I looked around my office
\item  I didn't see anyone nearby
\item  So, I went to the park during my break
\item  and found a quiet spot there
\item  That feeling came back again
\item  I looked around once more
\item  I still didn't see anyone nearby
\item  but I decided to take a chance
\item  I wasn't too far from my office, maybe a ten-minute walk
\item  I found a bench to sit on and closed my eyes – uh – uh –
\item  that feeling came back once more
\item  I just took a deep breath
\item  I was sitting under a big oak tree now
\item  I looked up at the branches above me
\item  Nobody was around
\item  And uh I said to myself, well now,
\item  I had been taught that prayers can make a difference
\item  And that it doesn't hurt to try
\item  I knew I wasn't perfect, but I wanted to help my dad
\item  I'd admit that to anyone
\item  But I was willing to give it a try
\item  I looked around one more time
\item  and then I closed my eyes
\item  and said a quiet prayer
\item  I whispered, "God, I don't know if you're listening
\item  But I ask you to please help my father recover
\item  So he can regain his independence and enjoy life again
\item  I opened my eyes
\item  and went back to work
\item  Suddenly, I received a text message from my father
\item  It said, "I just took a few steps with my walker
\item  I – I couldn't believe it
\item  It felt like a sign that my prayer had been heard
\item  I rushed home after work
\item  and saw my father standing in the living room with the help of his walker
\item  He hadn't been able to do that in months without assistance
\item  And there he was, taking small steps
\item  That was the turning point in his recovery
\item  He continued to improve
\item  and regained most of his mobility within a year
\item  It's been, what, six or seven years since then?
\item  My father is now able to walk without any assistance
\item  He's living life to the fullest, even in his seventies
\end{enumerate}\end{story_example}

\begin{story_example}[ex:hester-v2-church]{Narrative hester-v2-church }
\begin{enumerate}[itemsep=0mm]
\item Let me share a story from when I was twenty-two years old, just fresh out of college.
\item I was staying in a small town about five miles from here.
\item I had just started my first job at a local bookstore.
\item It was a quiet afternoon, and I was alone in the store.
\item My grandpa, who had been bedridden for almost a year, was at home with my sister.
\item She had to care for him daily, as he couldn't move around on his own.
\item There were only two of us siblings, my sister and me.
\item My parents had gone out of town for the weekend.
\item As I was shelving books, I suddenly heard a voice inside my head.
\item It said, "If you go to the old church down the street and light a candle for your grandpa, he will get better."
\item I looked around the bookstore, but no one was there.
\item I continued with my work, trying to ignore the voice.
\item It spoke to me again, repeating the same message.
\item I looked around once more, but still found no one.
\item I decided to take a break and walk to the church.
\item I was only a short distance away, maybe a quarter-mile.
\item As I approached the church, I hesitated for a moment.
\item I was not a particularly religious person.
\item I remembered my grandma always saying that prayers from a sincere heart could bring miracles.
\item On the other hand, she also said that prayers without faith were useless.
\item I admitted to myself that I was far from being a saint.
\item I had my share of mistakes and misdeeds.
\item I thought about my grandpa and how much he meant to our family.
\item Finally, I went inside the church and found a candle.
\item I knelt down in front of the altar.
\item I said, "God, I don't know how to pray properly.
\item But I am asking you with all my heart to heal my grandpa.
\item Please help him regain his strength and guide us in taking care of him."
\item After saying my prayer, I returned to the bookstore.
\item I resumed my work, shelving books.
\item Suddenly, I felt a strong urge to go home.
\item It was as if something was pulling me towards the house.
\item I didn't hear any voice this time.
\item It was just an overwhelming feeling that I needed to go.
\item I locked up the store and rushed home.
\item To my surprise, I found my grandpa sitting in the living room, talking to my sister.
\item He hadn't been able to leave his bed for months without assistance.
\item My sister told me that he had suddenly felt a surge of energy and was able to stand up and walk.
\item That day marked the beginning of his recovery.
\item He gradually regained his strength and independence.
\item It was a turning point in our lives.
\item We became more grateful for the time we had together.
\item I often thought back to that day in the church.
\item I wondered if my prayer had anything to do with my grandpa's recovery.
\item I couldn't be sure, but it definitely changed my perspective on faith and the power of prayer.
\item My grandpa lived for many more years after that incident.
\item He celebrated his ninetieth birthday surrounded by family and friends.
\item We cherished every moment we had with him.
\item That experience happened almost twenty years ago.
\item It remains a vivid memory in my mind.
\item My grandpa passed away a few years later, but his spirit lives on in our hearts.
\item Sometimes, when I'm feeling lost or uncertain, I think back to that day.
\item I remember the power of faith and the miracles it can bring.
\item And I am reminded that we are never truly alone, even when it seems like we are.
\end{enumerate}\end{story_example}


\newpage

Below, \ref{ex:boyscout} is taken directly from \citep{Labov1966}. \ref{ex:boyscout-scrambled} is the scrambled version used in experiments.

\begin{multicols}{2}

\begin{story_example}[ex:boyscout]{Narrative boyscout }
\begin{enumerate}[leftmargin = *,itemsep=0mm]
\item Yeah, I was in the boy scouts at the time. 
\item And we was doing the 50-yard dash, 
\item racing, 
\item but we was at the pier, marked off, 
\item and so we was doing the 50-yard dash. 
\item There was about 8 or 9 of us, you know, going down, coming back. 
\item And, going down the third time, I caught cramps
\item and I started yelling "Help!", 
\item but the fellows didn't believe me, you know. 
\item They thought I was just trying to catch up, because I was going on or slowing down. 
\item So all of them kept going. 
\item They leave me. 
\item And so I started going down. 
\item Scoutmaster was up there. 
\item He was watching me. 
\item But he didn't pay me no attention either. 
\item And for no reason at all there was another guy, who had just walked up that minute...
\item He just jumped over 
\item and grabbed me.
\end{enumerate}\end{story_example}

\columnbreak

\begin{story_example}[ex:boyscout-scrambled]{Narrative boyscout-scrambled }
\begin{enumerate}[leftmargin = *,itemsep=0mm]
\item They leave me. 
\item racing, 
\item Yeah, I was in the boy scouts at the time. 
\item but we was at the pier, marked off, 
\item and so we was doing the 50-yard dash. 
\item And we was doing the 50-yard dash, 
\item but the fellows didn't believe me, you know. 
\item and I started yelling "Help!", 
\item and grabbed me.
\item Scoutmaster was up there. 
\item So all of them kept going. 
\item And so I started going down. 
\item There was about 8 or 9 of us, you know, going down, coming back. 
\item He just jumped over 
\item They thought I was just trying to catch up, because I was going on or slowing down. 
\item And, going down the third time, I caught cramps
\item And for no reason at all there was another guy, who had just walked up that minute...
\item He was watching me. 
\item But he didn't pay me no attention either. 
\end{enumerate}\end{story_example}

\end{multicols}

\newpage

  
\ref{ex:triplett-v1-rookie-scrambled} and \ref{ex:hester-v2-church-scrambled} are the scrambled versions of \ref{ex:triplett-v1-rookie} and \ref{ex:hester-v2-church}, respectively. We do not reproduce here the scrambled version of the longest story ($L = 130$). These stimuli can be found in our code and data repository. 

The scrambling is done by taking a random permutation of the clauses. The numbering in the scrambled versions reflects the presentation order. 
\begin{story_example}[ex:triplett-v1-rookie-scrambled]{Narrative triplett-v1-rookie-scrambled }
\begin{enumerate}[itemsep=0mm]
\item and we traced down a little more
\item and left.
\item And I did.
\item I never observed it.
\item I was strategizing.
\item Big intimidating look on his face.
\item and proposed “Well we'll search for her
\item Back in the days when I was a rookie in the police force, the Chief was a veteran.
\item and we couldn't locate her.
\item And in those days I was a fitness enthusiast.
\item And we went to where they found her scarf – near a park.
\item But – nevertheless – that resolved the issue for that day.
\item That was the end of my rookie year.
\item And it seemed she was trying to impress me.
\item or had been awarded for bravery.
\item He had caught a notorious criminal –
\item She hadn't run away.
\item She was a chatterbox.
\item I managed to calm him down.
\item – it was a police station –
\item And returned
\item Truth is I wasn't very fond of her because she was –
\item Then she left a message one day saying she was going to run away because he was always scolding her about me.
\item But that evening the deputy, Frank Mitchell, said “You better transfer and leave because that old man never forgets anything once he gets it into his mind.”
\item I transferred.
\item But he had – a young daughter.
\item Goodness, she could talk.
\item So he accepted my offer.
\item She was a charming girl until she opened her mouth.
\item she was sitting on a chair with a book in her hands.
\item He arrived at my apartment.
\item and if we can't locate her well you can – do whatever you think is right.”
\end{enumerate}\end{story_example}


\begin{story_example}[ex:hester-v2-church-scrambled]{Narrative hester-v2-church-scrambled }
\begin{enumerate}[itemsep=0mm]
\item  I remembered my grandma always saying that prayers from a sincere heart could bring miracles
\item  Finally, I went inside the church and found a candle
\item  As I was shelving books, I suddenly heard a voice inside my head
 
\item  I remember the power of faith and the miracles it can bring
\item  It was a turning point in our lives
\item  That experience happened almost twenty years ago
\item  I was only a short distance away, maybe a quarter-mile
\item  My grandpa lived for many more years after that incident
\item  I said, "God, I don't know how to pray properly
\item  And I am reminded that we are never truly alone, even when it seems like we are
\item  It spoke to me again, repeating the same message
\item  I was not a particularly religious person
 
\item  Suddenly, I felt a strong urge to go home
\item  I looked around the bookstore, but no one was there
\item  I was staying in a small town about five miles from here
\item  My sister told me that he had suddenly felt a surge of energy and was able to stand up and walk
\item  Sometimes, when I'm feeling lost or uncertain, I think back to that day
\item  I wondered if my prayer had anything to do with my grandpa's recovery
 
\item  Let me share a story from when I was twenty-two years old, just fresh out of college
\item  My parents had gone out of town for the weekend
\item  I locked up the store and rushed home
\item  It was as if something was pulling me towards the house
\item  I admitted to myself that I was far from being a saint
\item  I had just started my first job at a local bookstore
\item  As I approached the church, I hesitated for a moment
\item  She had to care for him daily, as he couldn't move around on his own
\item  It was a quiet afternoon, and I was alone in the store
\item  We cherished every moment we had with him
 
\item  My grandpa, who had been bedridden for almost a year, was at home with my sister
\item  I had my share of mistakes and misdeeds
\item  He hadn't been able to leave his bed for months without assistance
 
\item  There were only two of us siblings, my sister and me
\item  I didn't hear any voice this time
\item  But I am asking you with all my heart to heal my grandpa
\item  My grandpa passed away a few years later, but his spirit lives on in our hearts
\item  I looked around once more, but still found no one
\item  He celebrated his ninetieth birthday surrounded by family and friends
\item  That day marked the beginning of his recovery
\item  We became more grateful for the time we had together
\item  I continued with my work, trying to ignore the voice
\item  It was just an overwhelming feeling that I needed to go
\item  I knelt down in front of the altar
\item  I thought about my grandpa and how much he meant to our family
\item  Please help him regain his strength and guide us in taking care of him
\item  To my surprise, I found my grandpa sitting in the living room, talking to my sister
\item  I often thought back to that day in the church
\item  I decided to take a break and walk to the church
\item  I couldn't be sure, but it definitely changed my perspective on faith and the power of prayer
\item  After saying my prayer, I returned to the bookstore
\end{enumerate}\end{story_example}

\subsection{List of Nouns}
\label{sec:list-of-nouns}

The following list of 32 nouns in exactly this order was presented to each participant in the Random List of Nouns experiment.

\begin{example}[ex:list-of-nouns]{}
'FAN', 'POLICE', 'SURVEY', 'COFFEE', 'ISLAND', 'CALF', 'ACID',
       'SUITCASE', 'TIDE', 'BULLET', 'BUILDING', 'TELEVISION', 'QUEEN',
       'BEDROOM', 'LIGHTNING', 'MOUNTAIN', 'LEMON', 'SEAT', 'CITY',
       'VICTIM', 'LIST', 'CHAIN', 'STEM', 'ATOM', 'STAKE', 'DEER',
       'OPERATOR', 'CROSS', 'FLOOR', 'INFANT', 'INDIAN', 'PASSAGE'
\end{example}

\subsection{Story used for LLM reliability assessment}
\label{sec:reliability}

The following narrative was used for comparison of recall performance evaluation by humans versus GPT in \cref{fig:reliability}.

The story was generated with an older version of GPT (\ff{text-davinci-003}) using the following prompt seeded by the boyscouts narrative  (\ref{ex:boyscout}, in paragraph form without the numbering):

\vspace{\baselineskip}

\begin{prompt}[ex:panic_gen]{Panic Attack Story Generation}
\code{This is an oral narrative about a life-threatening event:\\
$\{\ff{story}\}$\\
Create a different, longer and more detailed oral narrative with a similar narration style about another life-threatening event.}
\end{prompt}

 yielding the following narrative:
\begin{argument}[ex:panic]{Panic Attack Narrative}

Well, I was at work one day and I was doing my usual job - working on the warehouse floor, stocking boxes and organizing merchandise. I had been working for a few hours and it was getting late. I had been feeling a bit strange, a bit lightheaded, but chalked it up to hunger and not having eaten yet that day. But then, as I was walking around the warehouse returning items to their proper places, I started to feel a strange tightness and pressure in my chest. I stopped what I was doing and leaned against a nearby wall for a bit, but the pressure only got stronger. And I could feel my heart pounding in my chest. Next thing I know, I started seeing spots and I could feel sweat dripping down my forehead and my arms. I tried to take some deep breaths, but I just couldn't catch my breath. I was really scared and didn't know what to do. Suddenly I heard a voice coming from across the warehouse. It was the manager, who had seen me and called out to me - it seems someone had noticed me and was checking to see if I was alright. I tried to yell back that something was wrong, but I couldn't find my voice. The manager started walking towards me and I could feel my world going dark. I started to drop to the ground, but luckily the manager got to me in time and caught me just before my legs gave out. He took me to the staff room and sat me down in a chair. He asked me questions about what was wrong and what happened. He got me a glass of cold water and told me to rest for a bit. After a few minutes, the pressure in my chest slowly eased up and I started to feel better. It turned out I had been having a panic attack, but I escaped from the situation with the help of the quick action of the manager. It was never a fun experience, but it really could have been much worse, and it taught me to check in with myself and make sure to properly rest and eat to avoid further episodes.
\end{argument}

The story was subsequently segmented using the prompt:

\begin{prompt}[ex:panic_seg_prompt]{Segmentation Prompt}
\code{This is the original story:\\
$\{$\ref{ex:panic}$\}$\\
Split the sentences of the original story into the maximum number of small pieces, such that each piece contains minimal information. }
\end{prompt}

The segmentation that comes up is the following:
\begin{completion}[ex:panic_seg]{Panic Attack Segmentation}
1. I was at work. \\
2. I was working on the warehouse floor. \\
3. I was stocking boxes and organizing merchandise. \\
4. I had been working for a few hours. \\
5. It was getting late. \\
6. I had been feeling a bit strange. \\
7. I was feeling lightheaded. \\
8. I chalked it up to hunger and not having eaten yet that day. \\
9. I was walking around the warehouse returning items to their proper places. \\
10. I started to feel a strange tightness and pressure in my chest. \\
11. I stopped what I was doing. \\
12. I leaned against a nearby wall. \\
13. The pressure only got stronger. \\
14. I could feel my heart pounding in my chest. \\
15. I started seeing spots. \\
16. I could feel sweat dripping down my forehead and my arms. \\
17. I tried to take some deep breaths. \\
18. I couldn't catch my breath. \\
19. I was really scared. \\
20. I didn't know what to do. \\
21. I heard a voice coming from across the warehouse. \\
22. It was the manager. \\
23. Someone had noticed me. \\
24. The manager was checking to see if I was alright. \\
25. I tried to yell back that something was wrong. \\
26. I couldn't find my voice. \\
27. The manager started walking towards me. \\
28. I could feel my world going dark. \\
29. I started to drop to the ground. \\
30. The manager got to me in time. \\
31. He caught me just before my legs gave out. \\
32. He took me to the staff room. \\
33. He sat me down in a chair. \\
34. He asked me questions about what was wrong and what happened. \\
35. He got me a glass of cold water. \\
36. He told me to rest for a bit. \\
37. The pressure in my chest slowly eased up. \\
38. I started to feel better. \\
39. It turned out I had been having a panic attack. \\
40. I escaped from the situation with the help of the quick action of the manager. \\
41. It was never a fun experience. \\
42. It really could have been much worse. \\
43. It taught me to check in with myself. \\
44. It taught me to properly rest and eat to avoid further episodes. 
\end{completion}

Note that the segmentation in this particular case is not verbatim, the clauses have been transformed to first person while retaining most important information. If a verbatim segmentation is desired however, it can be achieved by utilizing the same prompt with \ff{gpt-4} although the number of clauses will be slightly different.

\section{Additional Results}

\subsection{Reliability of LLM scoring}
\label{sec:LLM_reliability}
Here we report additional details of the reliability for LLM narrative scoring.

In scoring the recalls, the human scorers (who are the authors on this paper) did not develop overly sophisticate scoring protocols. The directions were intuitive, essentially boiling down to ``did this clause from the original story appear in the recalled story". Despite the ambiguity of this procedure, there was very high inter-scorer correlation.

\begin{table}[ht]
	\caption{A comparison between gpt-4, mean human ($\bar{h}$), and individual human scorers ($h_{1}$, $h_{2}$, $h_{3}$). Each entry in the table gives the correlation coefficient (r-value) between the corresponding row and column label. }
	\centering
\begin{tabular}{c|ccccc}
   r &    $\bar{h}$   &   $h_{1}$ & $h_{2}$ & $h_{3}$          \\ \hline
{\rm gpt-4-0613}  & 0.94 & 0.92 & 0.90 & 0.90\\
$\bar{ h}$ &        &  0.95 & 0.98 & 0.97  \\
$h_{1}$  &       &    &  0.89 & 0.86 \\
$h_{2}$  &       &    &   & 0.96
\end{tabular}
\label{table:scoring}
\end{table}

\subsubsection{Drift in Performance over time}\label{sec:drift}

The models offered through the OpenAI API are known to undergo regular updates. We found that within the span of two months, the correlation coefficient with mean human performance rose by $0.05$. We compare \ff{gpt-4} called via OpenAI's API on 05/23/23, the same model \ff{gpt-4} called less than two months later on 07/03/23, with the deprecated model \ff{gpt-4-0613} (called after 06/13/23). Fortunately, the frozen model \ff{gpt-4-0613}, which we use throughout the paper, is essentially as good as the most recent version.

\begin{table}[ht]
	\caption{A comparison on gpt-4 scoring compiled at different times of the year. Evidently, the model improves over time in approximating the mean human scoring, with the most recent model achieving a correlation closest to one. The entries of the table show correlation coefficients (r-values) between the corresponding row and column variables.  }
	\centering
\begin{tabular}{c|ccc}
   r &  {\rm gpt-4} (05/23/23)  & {\rm gpt-4-0613}   &   {\rm gpt-4} (07/03/23)          \\ \hline
   $\bar{ h}$ &   0.89    &  0.937 & 0.944  \\
{\rm gpt-4} (05/23/23)  &       &  0.94 &  0.92  \\
{\rm gpt-4-0613}  &       &    &  0.99 
\end{tabular}
\label{table:correlations_drift_2}
\end{table}

\subsubsection{Comparison between different LLMs}
\label{sec:different_LLM}

\begin{figure}[htb!]
	\centering
 \includegraphics[width = \textwidth]{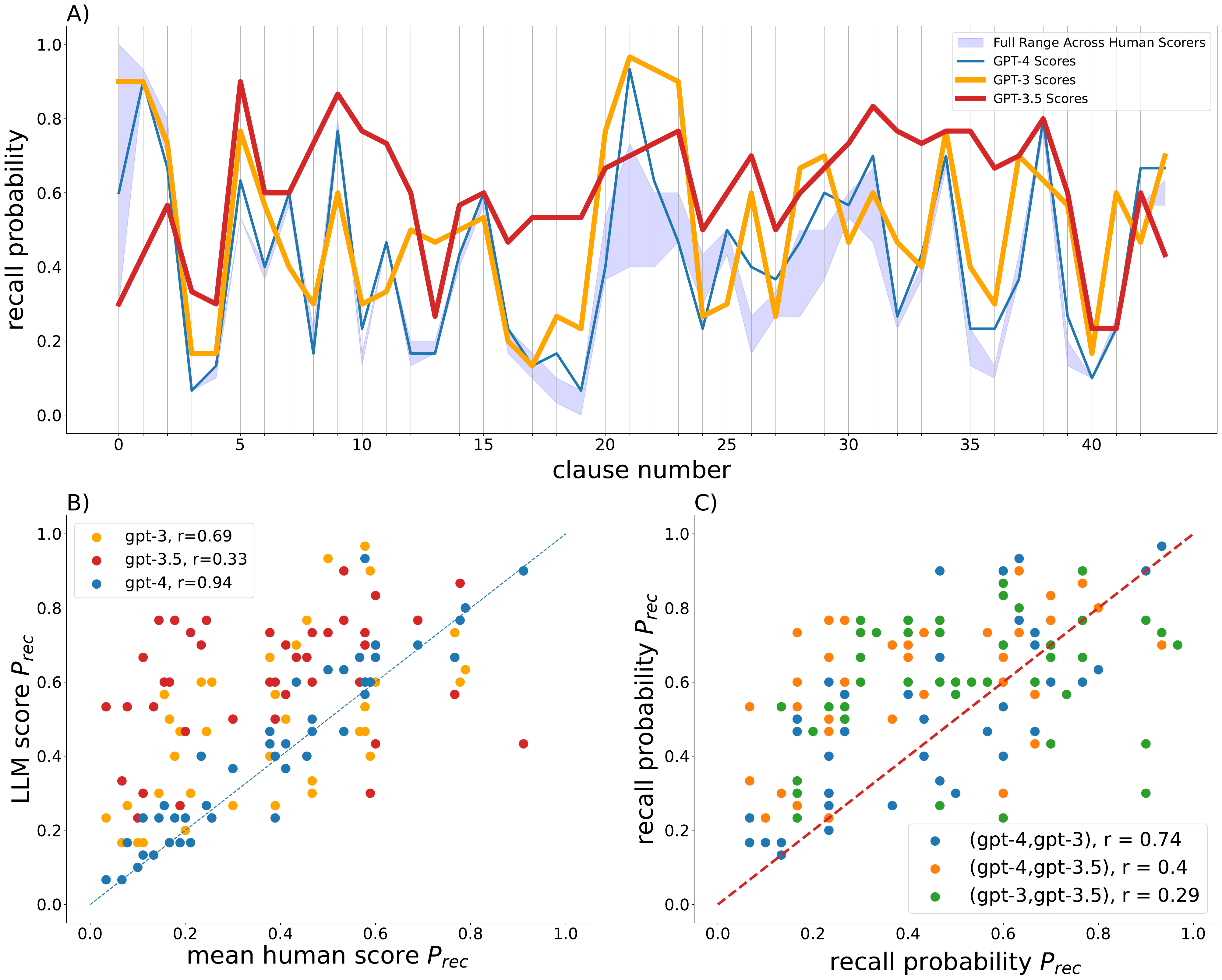}
	\caption{{\it Reliability of different LLMs for scoring narratives} We observed that performance on human recall scoring was not monotonic in the size of the LLM. (A) shows $P_{rec}$ for each clause as computed by LLMs of three different sizes, in order of increasing size: GPT-3 (OpenAI's API model \ff{text-davinci-003}), GPT-3.5 (or ChatGPT, model \ff{gpt-3.5-turbo}), and GPT-4 (\ff{gpt-4-0613}). GPT-3.5 scores appear systematically biased to be higher than the mean human scores, whereas GPT-3 roughly follows the human trend. (B) shows correlations between these models and mean human performance. (C) shows inter-LLM correlations. }
	\label{fig:cross_comparison_LLM}
\end{figure}

We compared different LLMs available through the OpenAI API. In order of increasing number of learned parameters (model size), these were GPT-3 (\ff{text-davinci-003}), GPT-3.5 (or ChatGPT, \ff{gpt-3.5-turbo}) and GPT-4 (\ff{gpt-4-0613}). The same scoring \ref{prompt:scoring} was used for all of these models. The results are visualized in \Cref{fig:cross_comparison_LLM}. We find that GPT-4 is qualitatively better than the smaller models (compare correlation between human and LLM scoring in \Cref{fig:cross_comparison_LLM}B). Furthermore, increasing size does not seem to be sufficient to increase performance, as illustrated by the severe drop in performance by GPT-3.5. As seen in \Cref{fig:cross_comparison_LLM}(A,B), GPT-3.5 appears to score more generously than the other models, resulting in a systematic upward bias of nearly all recall probabilities. It is possible that this reflects the sensitivity of these models to prompts, and that with appropriate tuning of the scoring prompt, the performance of GPT-3.5 could be improved.

\clearpage




\subsection{Compressed Recall in Longer Stories}\label{app:compression}

\begin{figure}[htbp!]
	\centering
 \includegraphics[width = \textwidth]{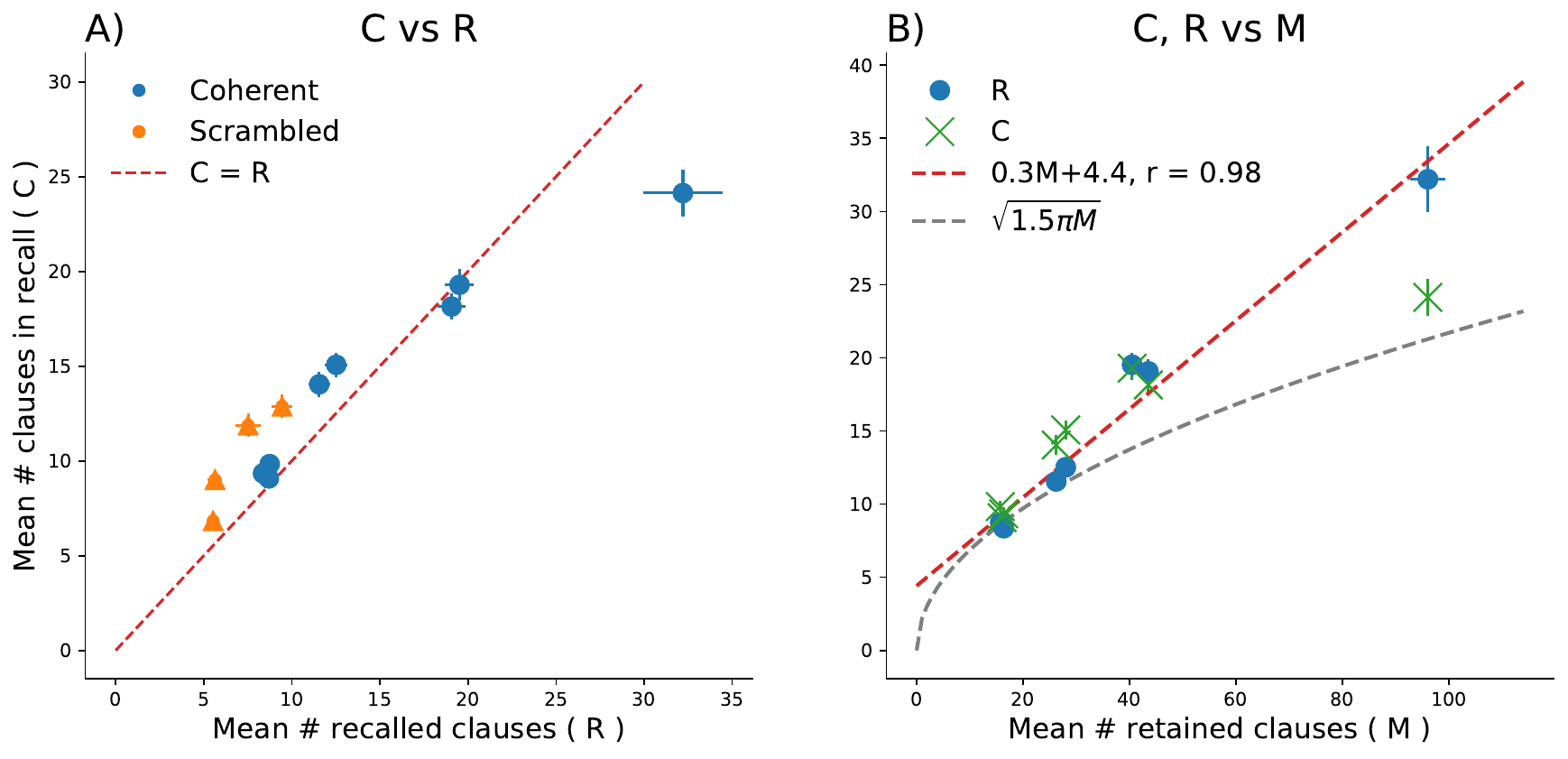}
 
\caption{ {\it Compression of Recalls}. Here we analyze how the length of recalls relates to story length and memory.
A) The vertical axis shows the mean number of clauses $C$ written for recall of each story, averaged over the participant population. $C$ for each recall was determined using the segmentation prompt given in this section. The horizontal axis gives $R$, the mean number of clauses from the narrative that were scored as being recalled. B) Plotting $R$ (blue dots) and $C$ (green crosses) vs M, as well as the linear fit (red dashed) to $(R,M)$. For the longest  narrative, $C$ deviates significantly from the linear fit to $R$, suggesting that a more conservative scoring of recalls would yield a sublinear scaling with $M$. 
}
	\label{fig:recall_compression}
\end{figure}

Each recall can be split up into a number of clauses. We denote the mean number of clauses used in recalls by $C$. \Cref{fig:recall_compression}A compares $C$ to the mean number of recalled clauses, defined previously and denoted by $R$. If every clause used by the participant were recalling a single clause from the original story, then the points would lie on a straight line, corresponding to $C = R$. For shorter stories, we find in fact that $C> R$, and people appear to use more clauses than necessary. However, for the longest story ($L = 130$), there is a marked compression, in which participants on average use $75\%$ as many clauses as they are scored to have recalled. It is possible that this reflects a tendency of GPT-4 to score generously when it encounters summaries or compressed descriptions in the recall. \Cref{fig:recall_compression}B shows how both $R$ and $C$ scale with the estimate for the mean number of encoded clauses $M$. The linear fit to $R$ is presented for comparison (also shown in \Cref{fig:RvM}), to illustrate the deviation of $C$ from linear scaling at longer stories. 

We used \ref{prompt:segmentation} with GPT-4 at $T = 0$ to segment recalls. An example of this applied to \ref{arg:boyscout_recall} is given in \ref{comp:seg_comp}.

\begin{prompt}[prompt:segmentation]
{Segmentation Prompt}
\code{Provide a word-for-word segmentation of the following narrative into linguistic clauses, numbered in order of appearance in the narrative:\\
\{\ff{narrative}\}
}

\end{prompt}

\begin{completion}[comp:seg_comp]{\ref{prompt:segmentation} using  \ref{arg:boyscout_recall}}
\code{1. A boy was with the scoutmaster  \\
2. and he was going up and down  \\
3. and on the third time of going up the pier  \\
4. he had cramps  \\
5. and shouted 'help'  \\
6. but no-one believed him,  \\
7. even the scoutmaster looked on  \\
8. by until this guy came out of nowhere  \\
9. and grabbed the boy  \\
10. and saved him. }
\end{completion}

\subsection{Lack of Output Interference in Recognition Experiments}
\label{app:output_interference}

Similar to \citep{criss2011output}, we measure the strength of output interference in recognition experiments by tracking the $d'$ measure as a function trial number. The discrimination measure used here is conventionally called simply $d'$ and is given by the difference in z-score between the hit probability $P_{h}$ (i.e. true positive rate) and the false alarm probability $P_{f}$ (i.e. false positive rate ). The z-transform is defined

\begin{align}
    z(p) = \sqrt{2} {\rm erf}^{-1}\left( 2 p-1\right),
\end{align}

and the discrimination measure is

\begin{align}
    d' = z(P_{h}) - z(P_{f}).
\end{align}

\begin{figure}[ht!]
	\centering
 \includegraphics[width = 1\textwidth]{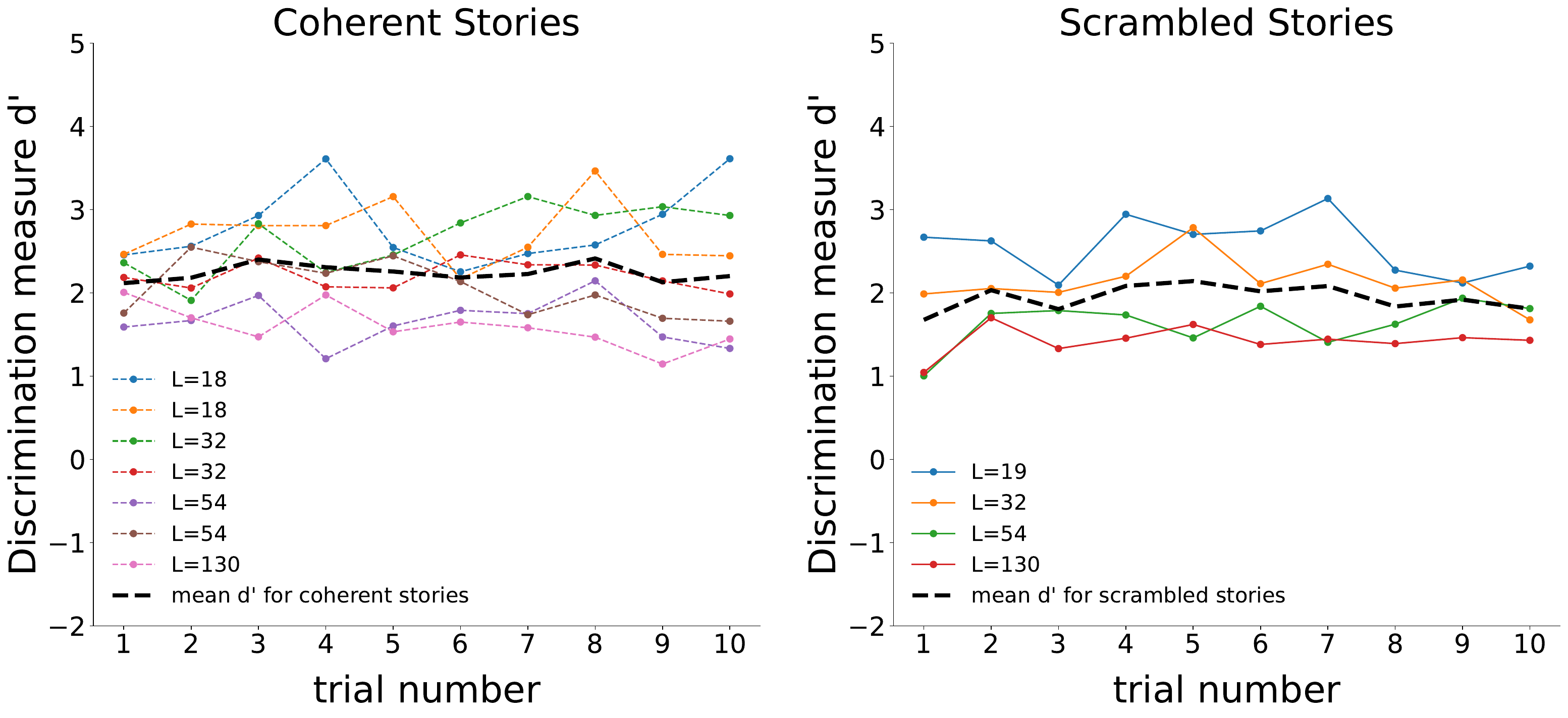}
	\caption{Discrimination measure shows good memory for true clauses which does not change significantly as the recognition trial progresses.  }
	\label{fig:d-prime}
\end{figure}

We plot $d'$ over the course of the recognition trials in \Cref{fig:d-prime} separately for coherent stories and scrambled stories. Output interference is characterized by a $d'$ which decreases with trial number, indicating a diminishing ability to discriminate new vs old items. 

\subsection{Recall Probability Curves and Distributions}
\label{app:serial-position}

\begin{figure}[!htb]
 \includegraphics[ width = 1\textwidth]{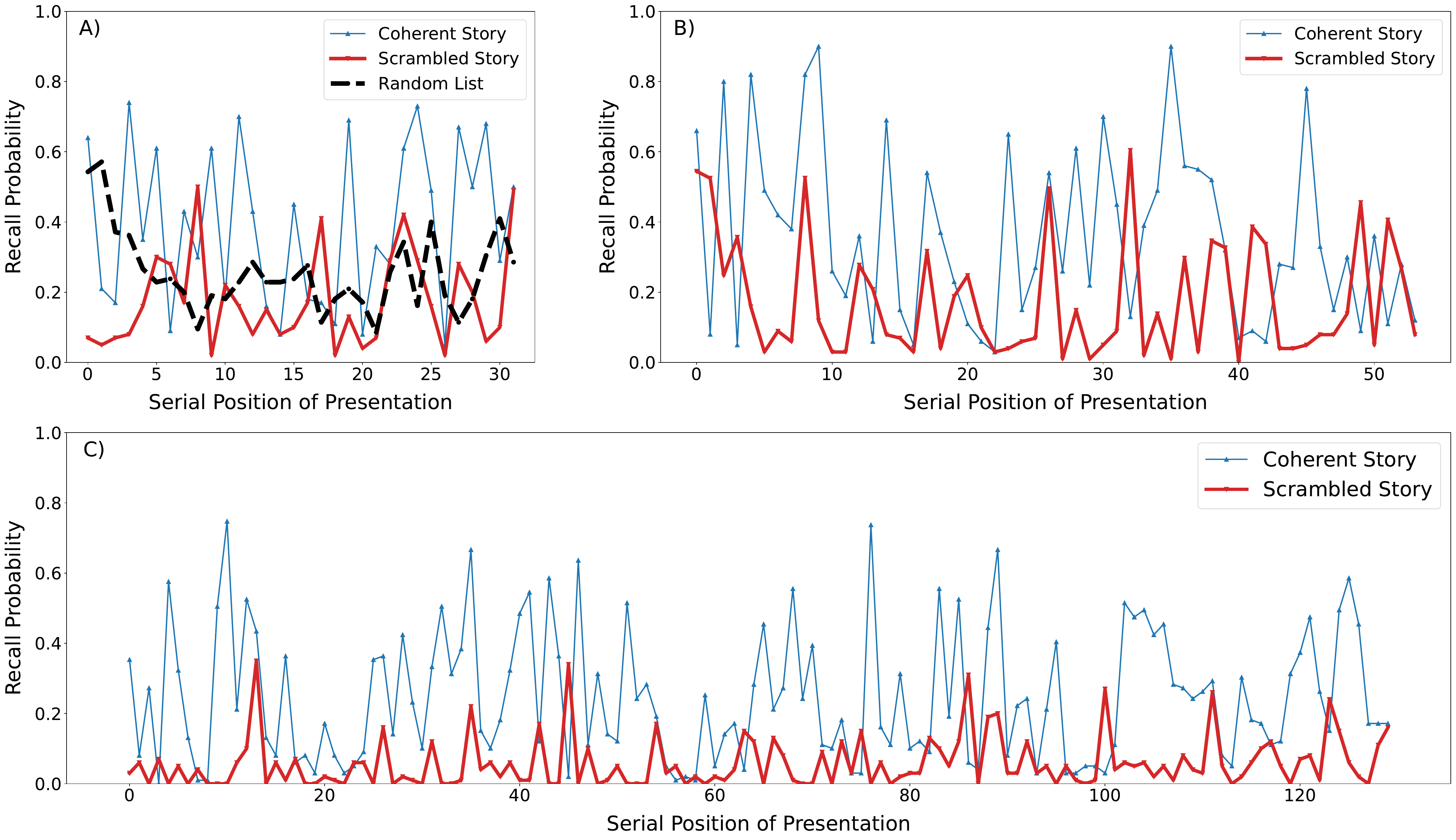}

	\caption{{\it Recall probability per item for narratives and random lists of words}. (A) shows the recall probability per clause as a function of the serial position of presentation for a coherent narrative (blue) and the scrambled version of this same narrative (red). As a comparison, we plot the recall probability per {\it word} as a function of serial position of presentation (black dashed). While the recall probability for the list shows a marked primacy effect, with a general suppression of recall in the bulk of the list, the recall of coherent narratives shows many spikes in $P_{rec}$ throughout the narrative. And while the scrambled stories have overall much lower recall probabilities, they still show large peaks within the bulk of the presentation. (B) and (C) compare coherent and scrambled narratives of increasing length.}
	\label{fig:serial}
\end{figure}

\begin{figure}[!htb]
 \includegraphics[width = 1\textwidth]{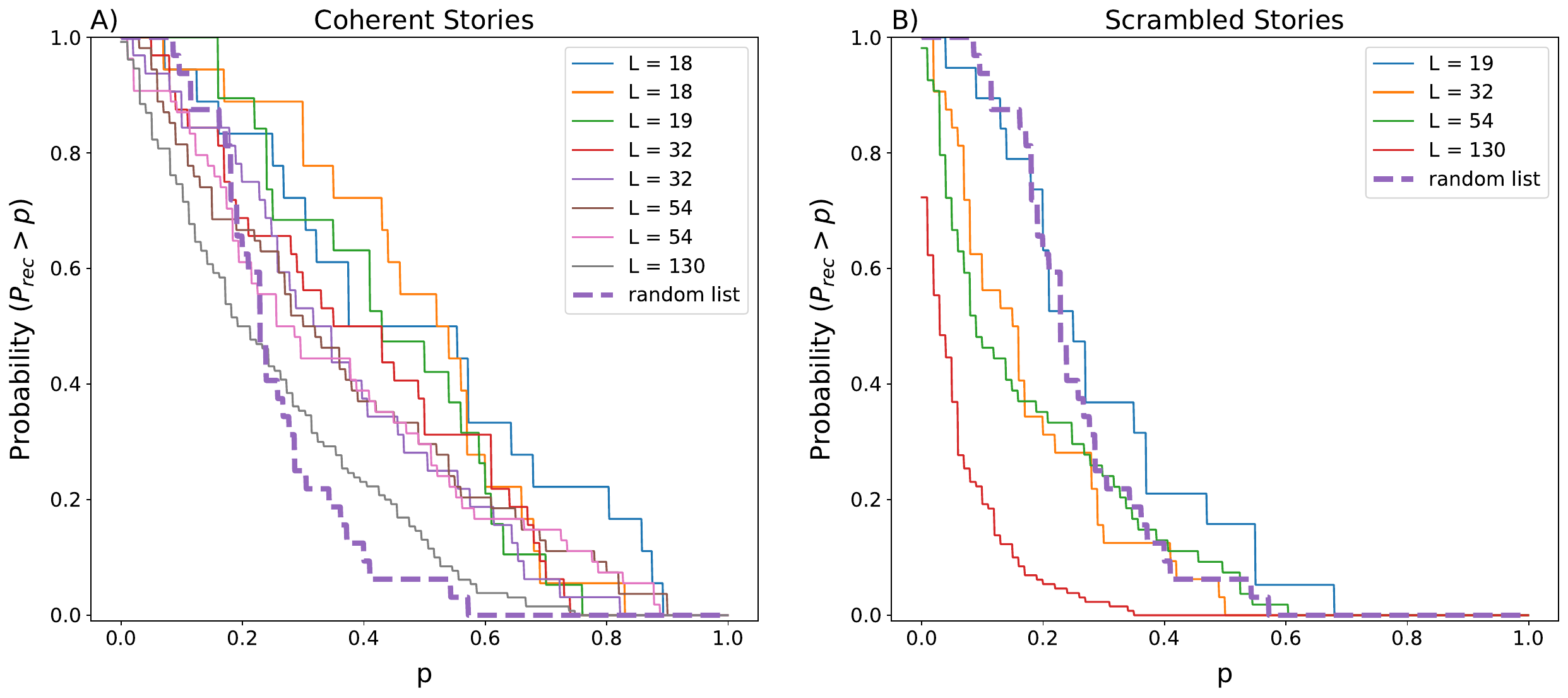}
	\caption{{\it Cumulative distribution of recall probabilities} (A) As compared to random lists (purple dashed), coherent narratives have a slower decline in the cumulative distribution function, indicating a larger number of clauses with a higher recall probability. (B) shows this behavior is reversed for scrambled stories, where the recall probability of clauses appears to uniformly drop, in some cases well below the corresponding distribution for recall of a random list. }
	\label{fig:prec_dist}
\end{figure}

Here we plot recall probability per clause as a function of the serial position of presentation. We refer to these below as recall probability curves. We compare recall probability curves for narratives to data obtained for random word lists. \Cref{fig:serial} a shows the recall probability curve for coherent (blue), scrambled (red) stories, alongside the recall probability curve for free recall of a random word list (black dashed). Importantly, for the scrambled recall, the serial position corresponds to the position of {\it presentation}. Two main points can be gleaned from these figures. First, whereas for random lists there is a pronounced primacy effect, we see no such structure in both coherent and scrambled stories. In fact, highly recalled clauses appear at all positions in the story. Another observation is that the distribution of $P_{rec}$ for clauses in a narrative covers a much broader range than $P_{rec}$ for words in a list. This can be seen by looking at the cumulative distribution function of $F(p) = P\left(P_{rec} > p\right)$, which is the probability to find $P_{rec} > p$. As shown in \Cref{fig:prec_dist}A, $F(p)$ drops to zero around $p \approx 0.6$ for random lists, while showing a more gradual decay for coherent narratives. In contrast, \Cref{fig:prec_dist}B shows that recall is significantly impaired for scrambled stories, with $F(p)$ showing a more precipitous fall to zero similar to random word lists.

\begin{figure}[!htb]
 \includegraphics[ width = 1\textwidth]{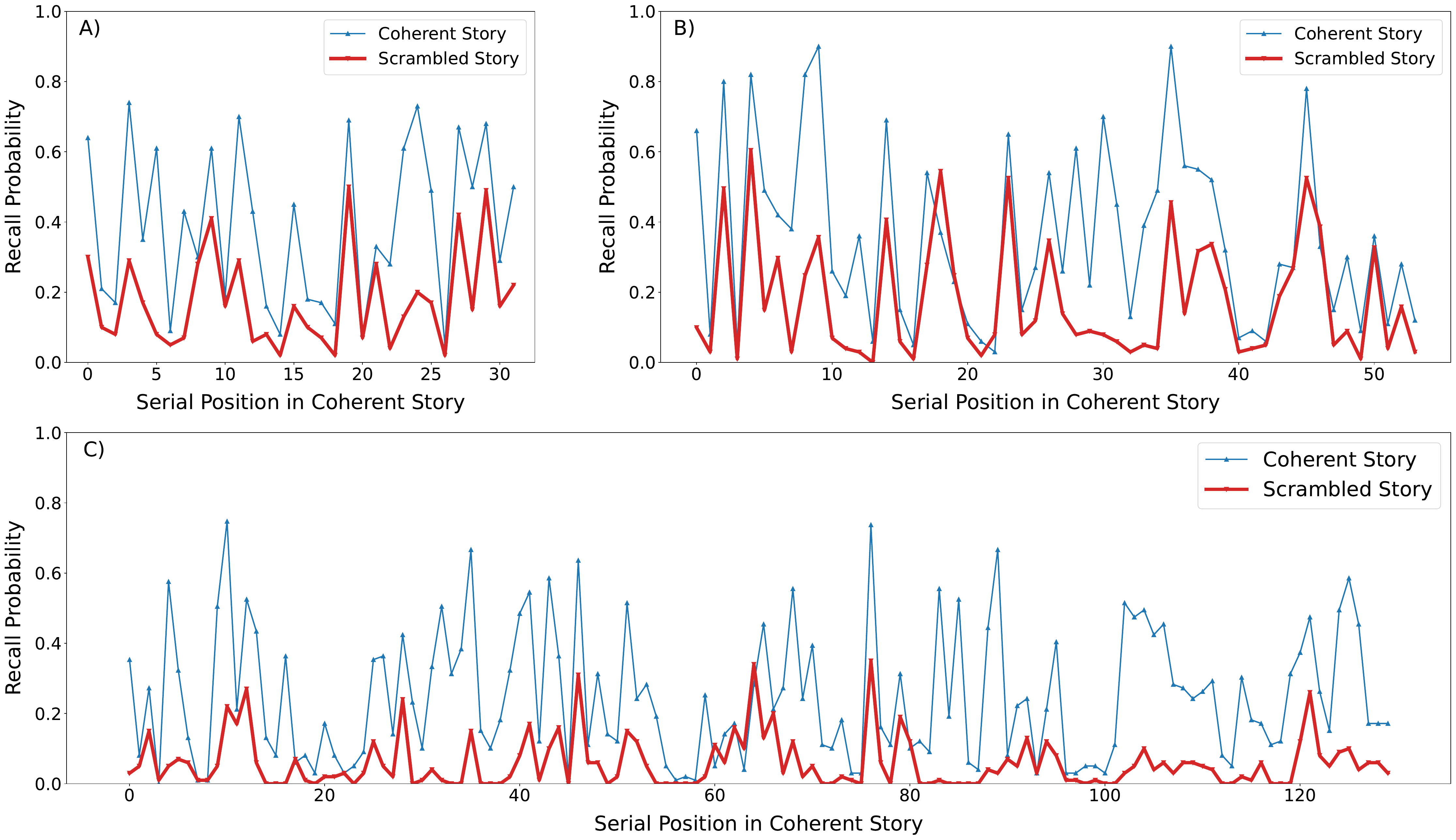}
 \includegraphics[width = 1\textwidth]{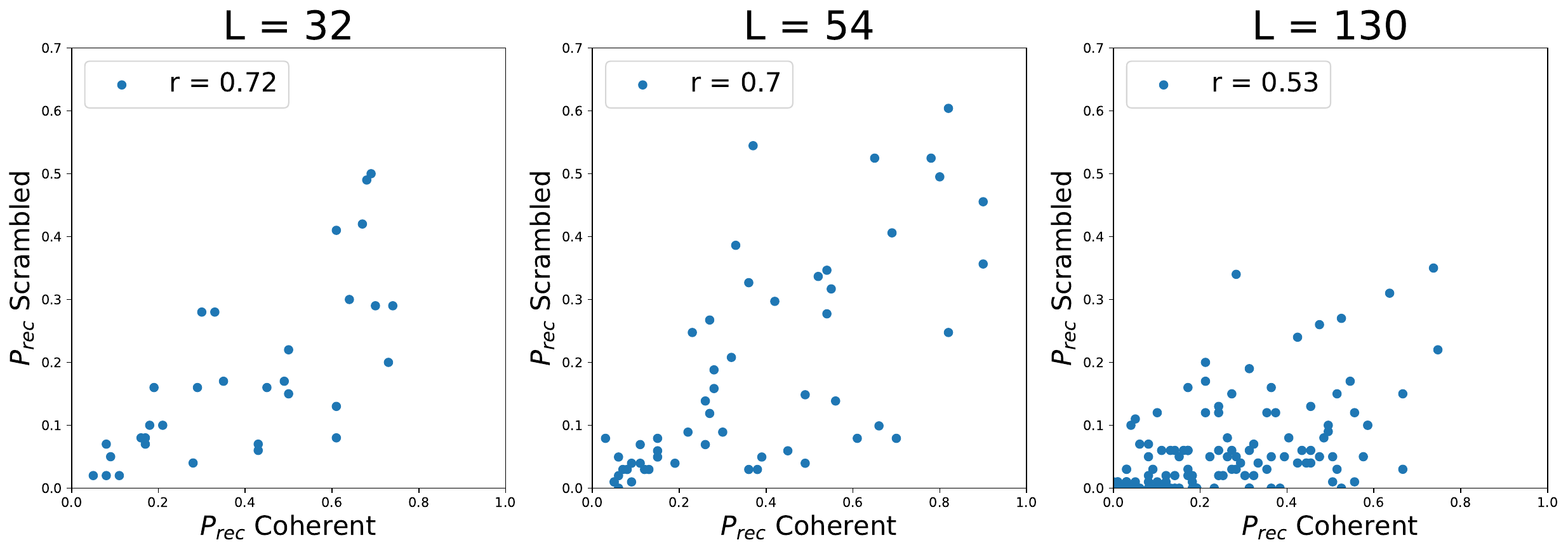}

	\caption{{Descrambling of narratives in recall}. (A-C) show recall probability as a function of the serial position in the coherent story. Note that for the scrambled $P_{rec}$ (red curves), this is {\it not} the presentation order, but rather the original order of the clauses. Plotting the curves in this way, we observe the spikes in recall probability appear to coincide between coherent and scrambled recall. Below, we show correlations between $P_{rec}$ for a given clause in the coherent and scrambled presentation.}
	\label{fig:serial-scramble}
\end{figure}

However, the recall probability curves for scrambled stories reveals a more subtle effect. While the overall distribution of recall probabilities may be similar to (if not worse than) random lists, there are still spikes in $P_{rec}$ throughout the bulk of the narrative, suggesting that these texts are not processed as if they were a random list of clauses. In \Cref{fig:serial-scramble}, we replot the recall of only the scrambled stories from \Cref{fig:serial} according the serial position of the clause {\it in the coherent story}. Below this, we show the correlation between recall probability of a clause appearing in the coherent story versus the scrambled version. While low, the correlations are positive and statistically significant, suggesting that participants are able to identify memorable clauses even in the scrambled scenario, and use this in attempting to construct a coherent recall. This active process of selection might explain why average recall of clauses for scrambled narratives appears {\it worse} than recall of words in random lists (i.e. why the scrambled narrative data (orange triangles) lie below the square-root scaling (dashed gray) in \Cref{fig:RvM}).

\clearpage

\subsection{Semantic Similarity and Recall}\label{sec:sim_appendix}

In the main text, we compared a measure of semantic similarity for each clause with its population averaged recall probability. The main text showed this for only a few narratives. Here, we display the data comparing this measure with recall probability for all narratives, as well as some summary figures. We also compare across a few different embedding models, some of which are publicly available. 

The text-to-vector embedding models take a block of text as input (no longer than 4096 tokens in length), and produce a real vector output, which we refer to as the semantic vector embedding of the text. The OpenAI API provides access to three text embedding models, each of which produces normalized vector embedding of dimension 1536 (\ff{text-embedding-3-small}, \ff{text-embedding-ada-002}) and 3072 (\ff{text-embedding-3-large}). We compare to an open-access embedding model from mixedbread.ai (\ff{mxbai-embed-large-v1}) which produces unnormalized 
vector embeddings of dimension 1024. 

The cosine similarity score we compute takes the vector embedding of a single clause ${\bf E}(c_{i})$, and computes the cosine similarity with the vector embedding of the entire narrative ${\bf E}(C)$, where $C = c_{1}c_{2}...c_{L}$. The cosine similarity score is

\begin{align}
    S_{i} = \frac{{\bf E}(c_{i}) \cdot {\bf E}(C)}{\Vert {\bf E}(c_{i})\rVert \, \lVert {\bf E}(C)\rVert},
\end{align}

where the numerator shows the dot-product of the two vectors, and $\lVert {\bf x} \rVert = \sqrt{ \sum_{i} x_{i}^{2}}$ denotes the norm of the vector ${\bf x}$.

\begin{figure}[!htb]

 \includegraphics[width = 1\textwidth]{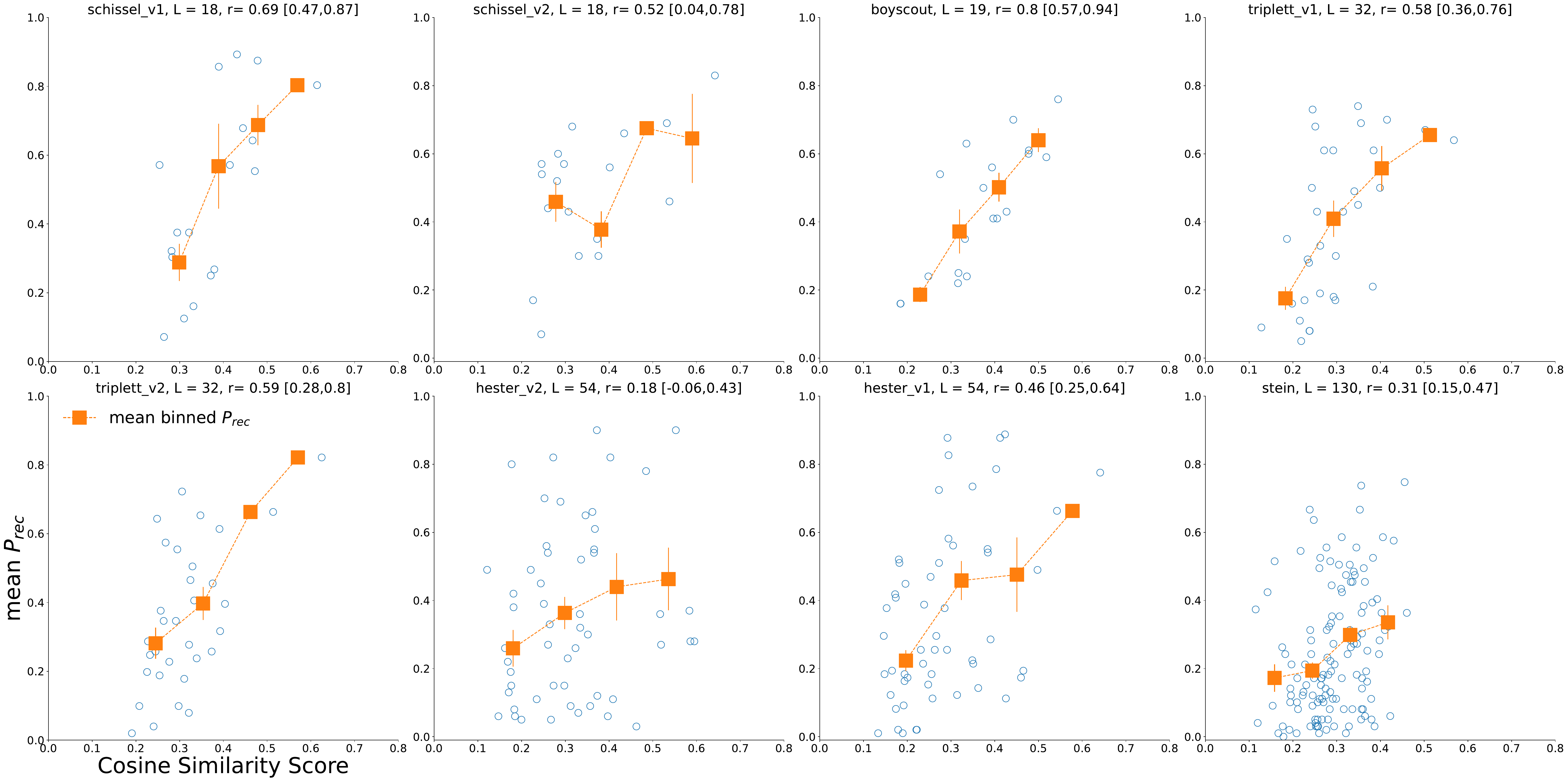}

	\caption{{\it Recall probability vs cosine similarity scores for all narratives} In each plot, we plot all $L$ points $(P_{rec}(c_{i}), S_{i})$  (blue circles). To compute the orange square points, we bin the full range of similarity scores into four equal intervals, and compute the average $P_{rec}$ in each interval. The horizontal coordinate for these points is taken to be the midpoint of the bin interval. The plots are labeled by the corresponding story name, along with the length $L$, and the correlation coefficient computed between $P_{rec}$ and the similarity score. The $95\%$ confidence interval for the correlation coefficient is indicated in brackets $[ r_{low}, r_{high}]$, and is computed with bootstrap using 1000 samples.}
	\label{fig:similarity-score-all-narratives}
\end{figure}

\begin{figure}[!htb]

 \includegraphics[width = 1\textwidth]{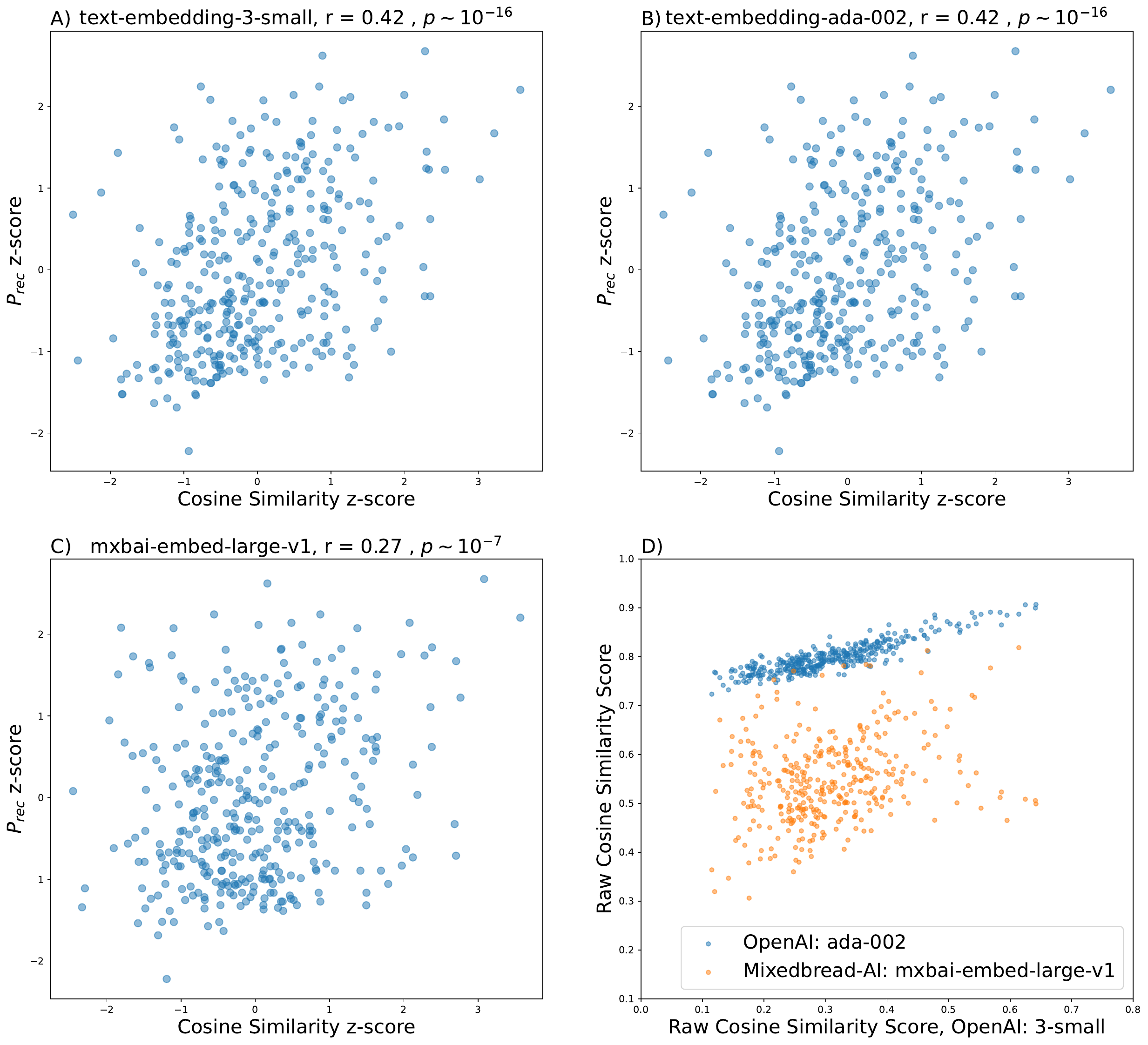}

	\caption{{\it Recall probability vs Similarity scores for all narratives} Here we compare different embedding models by calculating z-scores for recall and similarity values and pooling all of the data. We see very strong correlation between the OpenAI models, and a weaker but still good correlation with the open-sourced model. Not shown here is the correlation between \ff{text-embedding-3-large} and $P_{rec}$, which is $r = 0.44$ at $p \sim 10^{-18}$, only marginally better than the smaller and more efficient models. p-value is computed using two-sided Wald test.  }
	\label{fig:similarity-z-scores}
\end{figure}

Note that in \Cref{fig:similarity-score-all-narratives}, each narrative has a different range of similarity scores. This likely has to do with the overall length of the narrative, though evidently length alone does not fully account for the variability. Furthermore, each narrative has a different range of recall probabilities. In order to compare across narratives, we compute z-scores of $P_{rec}$ and $S$ for a given narrative, and combine these z-scores across narratives in \Cref{fig:similarity-z-scores}. There is an overall strong correlation which survives when averaging across narratives. Using z-scores also allows a meaningful comparison between different models, and we show two OpenAI models, along with an open access model from mixedbread-AI. A comparison of raw similarity scores is shown in \Cref{fig:similarity-z-scores}D, where a model's particular bias is evident. For instance, \ff{text-embedding-ada-002} tends to produce similarity scores always above $0.7$, indicating a strong clustering of vector embeddings inside a cone. The Mixedbread-AI embeddings produce a much larger range of embedding directions, but also show a weaker correlation with OpenAI's '3-small' embeddings. 

\begin{figure}[!htb]

 \includegraphics[width = 1\textwidth]{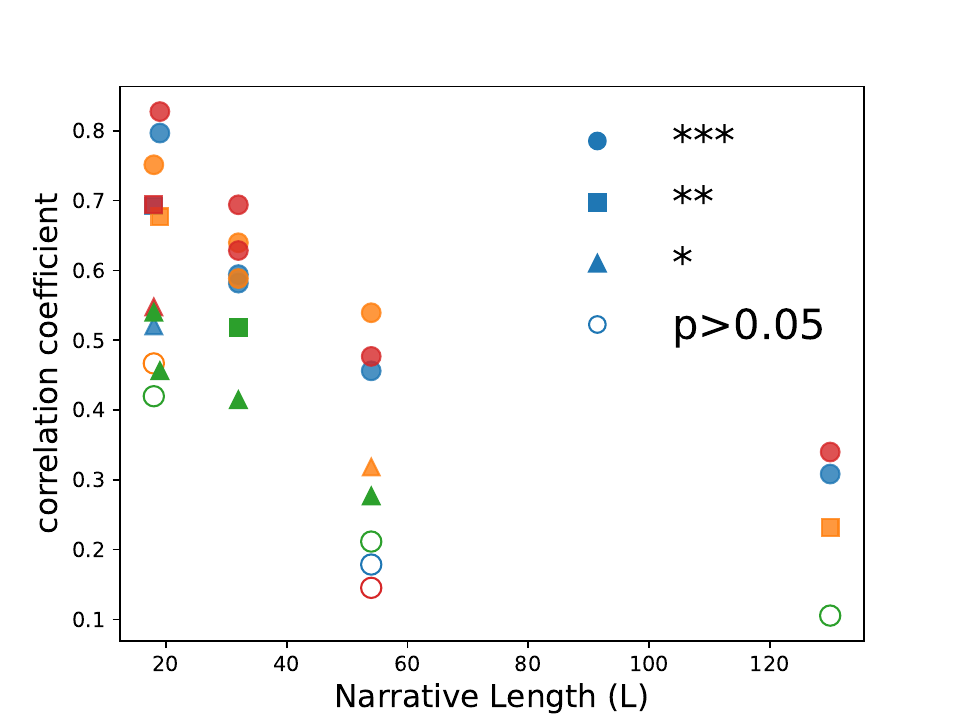}

	\caption{{\it Embedding model comparison} Here we show the correlation coefficient and statistical significance for each model on each narrative separately. The colors code for the embedding model: blue (OpenAI {\rm text-embedding-3-small}), orange (OpenAI {\rm text-embedding-ada-002}) and green (mixedbread.ai {\rm mxbai-embed-large-v1}). p-value is computed using two-sided Wald test.}
	\label{fig:r-vs-L-comparison}
\end{figure}

Finally, in \Cref{fig:r-vs-L-comparison}, we compute the correlation coefficient between recall probability and similarity score for each narrative individually (as in \Cref{fig:similarity_scores}), and compare across different LLM embedding models. All models show the same trend of decreasing correlation with increasing narrative length, discussed in the main text. Overall, the OpenAI models achieve higher correlation with lower p-value.

\clearpage

\subsection{Clauses with High Recall Probability}
\label{sec:a_hprc}
We noted that clauses recalled with higher probability tend to encapsulate a summarized version of the presented narrative. While not all nuances are retained, they collectively outline the principal storyline. Presented below are examples from stories of varying lengths. It's important to note that we present the recalled clauses exactly as they appeared in the original story and maintain the same sequence.

\newcounter{boxlblcounter}  
\newcommand{\makeboxlabel}[1]{$P_{rec}=#1$\hfill}
\newenvironment{plabel}
  {\begin{list}
    {\arabic{boxlblcounter}}
    {\usecounter{boxlblcounter}
     \setlength{\labelwidth}{0.5cm}
     \setlength{\labelsep}{0.5cm}
     \setlength{\itemsep}{0.0cm}
     \setlength{\leftmargin}{3.5cm}
     \setlength{\rightmargin}{0.5cm}
     \setlength{\itemindent}{-2cm} 
     \let\makelabel=\makeboxlabel
    }
  }
{\end{list}}

\begin{story_example}[ex:schissel_v1-pool]{Highly recalled clauses for narrative of length 18 clauses}

Original Story

\begin{enumerate}[label*=\arabic*.]
\item  My best friend pushed me into the pool.
\item  This happened during my cousin's wedding reception,
\item  where everyone was dressed to impress.
\item  And the reason it happened,
\item  she spotted a bee hovering near my face,
\item  - this was at a fancy hotel garden -
\item  and she tried to save me from being stung.
\item  And my aunt had just handed me a glass of champagne,
\item  and I warned her to be careful.
\item  'Course friends, y'know, they don't always think things through.
\item  So that's when she gave me a little shove,
\item  and I tumbled into the water.
\item  When I resurfaced, gasping for air,
\item  she just started laughing,
\item  and she apologized profusely.
\item  And... my beautiful dress was ruined,
\item  and naturally, the first thing to do was to get out and dry off,
\item  and my cousin just says, "Just about a few inches more," she says, "and you'd have landed on the cake."
\end{enumerate}

Highly recalled clauses
\begin{plabel}
\item[ 0.857]  1. My best friend pushed me into the pool.
\item[ 0.893]  2. This happened during my cousin's wedding reception,
\item[ 0.893]  5. she spotted a bee hovering near my face,
\item[ 0.661]  16. And... my beautiful dress was ruined,
\item[ 0.857]  18. and my cousin just says, "Just about a few inches more," she says, "and you'd have landed on the cake."
\end{plabel}

\end{story_example}

\begin{story_example}[ex:shester_v2-church]{Highly recalled clauses for narrative of length 54 clauses}

Original Story

\begin{enumerate}
\item  Let me share a story from when I was twenty-two years old, just fresh out of college
\item  I was staying in a small town about five miles from here
\item  I had just started my first job at a local bookstore
\item  It was a quiet afternoon, and I was alone in the store
\item  My grandpa, who had been bedridden for almost a year, was at home with my sister
\item  She had to care for him daily, as he couldn't move around on his own
\item  There were only two of us siblings, my sister and me
\item  My parents had gone out of town for the weekend
\item  As I was shelving books, I suddenly heard a voice inside my head
\item  It said, "If you go to the old church down the street and light a candle for your grandpa, he will get better
\item  I looked around the bookstore, but no one was there
\item  I continued with my work, trying to ignore the voice
\item  It spoke to me again, repeating the same message
\item  I looked around once more, but still found no one
\item  I decided to take a break and walk to the church
\item  I was only a short distance away, maybe a quarter-mile
\item  As I approached the church, I hesitated for a moment
\item  I was not a particularly religious person
\item  I remembered my grandma always saying that prayers from a sincere heart could bring miracles
\item  On the other hand, she also said that prayers without faith were useless
\item  I admitted to myself that I was far from being a saint
\item  I had my share of mistakes and misdeeds
\item  I thought about my grandpa and how much he meant to our family
\item  Finally, I went inside the church and found a candle
\item  I knelt down in front of the altar
\item  I said, "God, I don't know how to pray properly
\item  But I am asking you with all my heart to heal my grandpa
\item  Please help him regain his strength and guide us in taking care of him
\item  After saying my prayer, I returned to the bookstore
\item  I resumed my work, shelving books
\item  Suddenly, I felt a strong urge to go home
\item  It was as if something was pulling me towards the house
\item  I didn't hear any voice this time
\item  It was just an overwhelming feeling that I needed to go
\item  I locked up the store and rushed home
\item  To my surprise, I found my grandpa sitting in the living room, talking to my sister
\item  He hadn't been able to leave his bed for months without assistance
\item  My sister told me that he had suddenly felt a surge of energy and was able to stand up and walk
\item  That day marked the beginning of his recovery
\item  He gradually regained his strength and independence
\item  It was a turning point in our lives
\item  We became more grateful for the time we had together
\item  I often thought back to that day in the church
\item  I wondered if my prayer had anything to do with my grandpa's recovery
\item  I couldn't be sure, but it definitely changed my perspective on faith and the power of prayer
\item  My grandpa lived for many more years after that incident
\item  He celebrated his ninetieth birthday surrounded by family and friends
\item  We cherished every moment we had with him
\item  That experience happened almost twenty years ago
\item  It remains a vivid memory in my mind
\item  My grandpa passed away a few years later, but his spirit lives on in our hearts
\item  Sometimes, when I'm feeling lost or uncertain, I think back to that day
\item  I remember the power of faith and the miracles it can bring
\item  And I am reminded that we are never truly alone, even when it seems like we are
\end{enumerate}

Highly recalled clauses
\begin{plabel}
\item[ 0.710]  1. Let me share a story from when I was twenty-two years old, just fresh out of college.
\item[ 0.830]  3. I had just started my first job at a local bookstore.
\item[ 0.850]  5. My grandpa, who had been bedridden for almost a year, was at home with my sister.
\item[ 0.860]  9. As I was shelving books, I suddenly heard a voice inside my head.
\item[ 0.920]  10. It said, "If you go to the old church down the street and light a candle for your grandpa, he will get better."
\item[ 0.720]  15. I decided to take a break and walk to the church.
\item[ 0.700]  31. Suddenly, I felt a strong urge to go home.
\item[ 0.900]  36. To my surprise, I found my grandpa sitting in the living room, talking to my sister.
\item[ 0.780]  46. My grandpa lived for many more years after that incident.
\end{plabel}

\end{story_example}

\end{document}